%% file: main.tex
\documentclass[pdflatex,sn-apa,iicol]{sn-jnl}% Default with double column layout

\usepackage{graphicx}%
\usepackage{multirow}%
\usepackage{amsmath,amssymb,amsfonts}%
\usepackage{amsthm}%
\usepackage{mathrsfs}%
\usepackage[title]{appendix}%
\usepackage{xcolor}%
\usepackage{textcomp}%
\usepackage{manyfoot}%
\usepackage{booktabs}%
\usepackage{algorithm}%
\usepackage{algorithmicx}%
\usepackage{algpseudocode}%
\usepackage{listings}%
\usepackage[capitalise]{cleveref}

\usepackage{microtype}
\usepackage{caption}
\usepackage{subcaption}
\usepackage{hyperref}
\hypersetup{
    colorlinks=true,
    linkcolor=red,
    citecolor=blue,
    urlcolor=blue
}
\usepackage{verbatim}
\usepackage{forest}
\usetikzlibrary{arrows.meta, shapes.geometric, calc, fit}
\usepackage{makecell}
\newcolumntype{H}{>{\setbox0=\hbox\bgroup}c<{\egroup}@{}}
\usepackage{gensymb}
\usepackage{fontawesome}
\usepackage{pifont}% http://ctan.org/pkg/pifont
\newcommand{\cmark}{\ding{51}}%
\newcommand{\xmark}{\ding{55}}%
\definecolor{firstcolor}{RGB}{252, 235, 193}%{236 168 169}
\newcommand{\colorfirsttext}[1]{\colorbox{firstcolor}{\textbf{#1}}}
\usepackage{soul}
\newcommand{\etal}{\textit{et al}. }
\theoremstyle{thmstyleone}%
\theoremstyle{thmstyletwo}%

\theoremstyle{thmstylethree}%
\newtheorem{definition}{Definition}%

\begin{document}

\title[Article Title]{Advances in 3D Neural Stylization: A Survey}


\author*[1]{\fnm{Yingshu} \sur{Chen}} \email{ychengw@connect.ust.hk}

\author[1]{\fnm{Guocheng} \sur{Shao}} \email{\{gshao,kcshum\}@connect.ust.hk}

\author[1]{\fnm{Ka Chun} \sur{Shum}} %\email{kcshum@connect.ust.hk}
 
\author[2]{\fnm{Binh-Son} \sur{Hua}} \email{binhson.hua@tcd.ie}

\author[1]{\fnm{Sai-Kit} \sur{Yeung}} \email{saikit@ust.hk}

% \affil*[1]{\orgdiv{Department}, \orgname{The Hong Kong University of Science and Technology}, \orgaddress{\street{Street}, \city{City}, \postcode{100190}, \state{Hong Kong}, \country{China}}}
\affil*[1]{\orgname{Hong Kong University of Science and Technology}}

\affil[2]{\orgname{Trinity College Dublin}}

% \affil[3]{\orgdiv{Department}, \orgname{Organization}, \orgaddress{\street{Street}, \city{City}, \postcode{610101}, \state{State}, \country{Country}}}

\abstract{Modern artificial intelligence offers a novel and transformative approach to creating digital art across diverse styles and modalities like images, videos and 3D data, unleashing the power of creativity and revolutionizing the way that we perceive and interact with visual content. This paper reports on recent advances in stylized 3D asset creation and manipulation with the expressive power of neural networks. We establish a taxonomy for neural stylization, considering crucial design choices such as scene representation, guidance data, optimization strategies, and output styles. Building on such taxonomy, our survey first revisits the background of neural stylization on 2D images, and then presents in-depth discussions on recent neural stylization methods for 3D data, accompanied by a benchmark evaluating selected mesh and neural field stylization methods. Based on the insights gained from the survey, we highlight the practical significance, open challenges, future research, and potential impacts of neural stylization, which facilitates researchers and practitioners to navigate the rapidly evolving landscape of 3D content creation using modern artificial intelligence.}




\maketitle

\begin{figure*}
    \includegraphics[width=\linewidth]{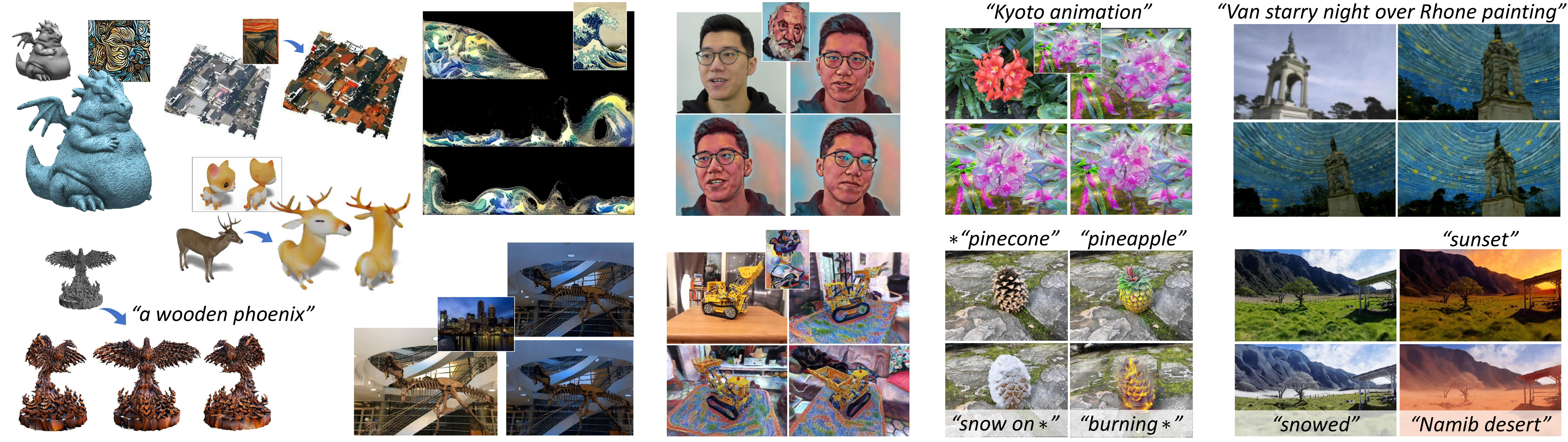}
    \centering
    \caption{The survey delves into the realm of neural stylization on diverse 3D representations, including meshes, point clouds, volume, and neural fields. The neural stylization with visual, textual and geometric features retrieved from large-scale neural models empowers artistic, photorealistic, and semantic style transformation of the geometry and appearance of 3D scenes. Images adapted from \cite{liu2018paparazzi, ma2023x, cao2020psnet, yin20213dstylenet, wang2023tsnerf, zhang2022arf, zhang2023ref, song2023blending, haque2023instructnerf}.}
    \label{fig:teaser}
\end{figure*}

\section{Introduction}\label{sec:intro}

% Background AI 2D and 3D
Digital art and visual design have been prevailing in our daily living spaces, expressing visually captivating aesthetics, unique tastes, and emotions of human beings.
With the prosperity of artificial intelligence (AI), there emerges a new generation of toolsets for visual content creation such as generative image synthesis (Stable Diffusion, DALL·E 3, Midjourney) \citep{rombach2022high, dalle3_2023, midjourney_2023}, and video synthesis (Sora, RunwayML) \citep{sora, runway_2023}. 
AI-based visual content creation can also be extended to the 3D domain, notably by lifting images to 3D scenes (LUMA AI) \citep{lumaai_2023}, and creative text-guided 3D generation and design (DreamFusion, Meshy, SplineAI) ~\citep{poole2022dreamfusion, meshy, spline_2023}. 

% Style 
It's noteworthy that the nature of visual concepts in our living space is tied to a critical factor: style.
Formally, style is a way to express individuality and creativity in different mediums and practices.
In relevant industries like animation, architectural and interior design, gaming, augmented reality, virtual reality, and artwork creation, assets are often created in styles such that they altogether harmonize to create an intended look and feel of the final scenes. 
A common practice is to first create the assets, and then post-process them to match some styles, often known as \emph{stylization}. 
The advent of modern deep learning has led to the emergence of \emph{neural stylization}, a family of methods that automatically create visual content in styles, facilitating the exploration of aesthetics in the creation of visual data. 
Neural stylization is applicable to visual data in general, including images, videos, and 3D data.
Unlike image stylization, which has been well developed in the past decade, neural stylization for 3D data remains an open area to explore for new creative vistas and practical applications.

\begin{figure}[tbp]
    \centering
    \includegraphics[width=\linewidth]{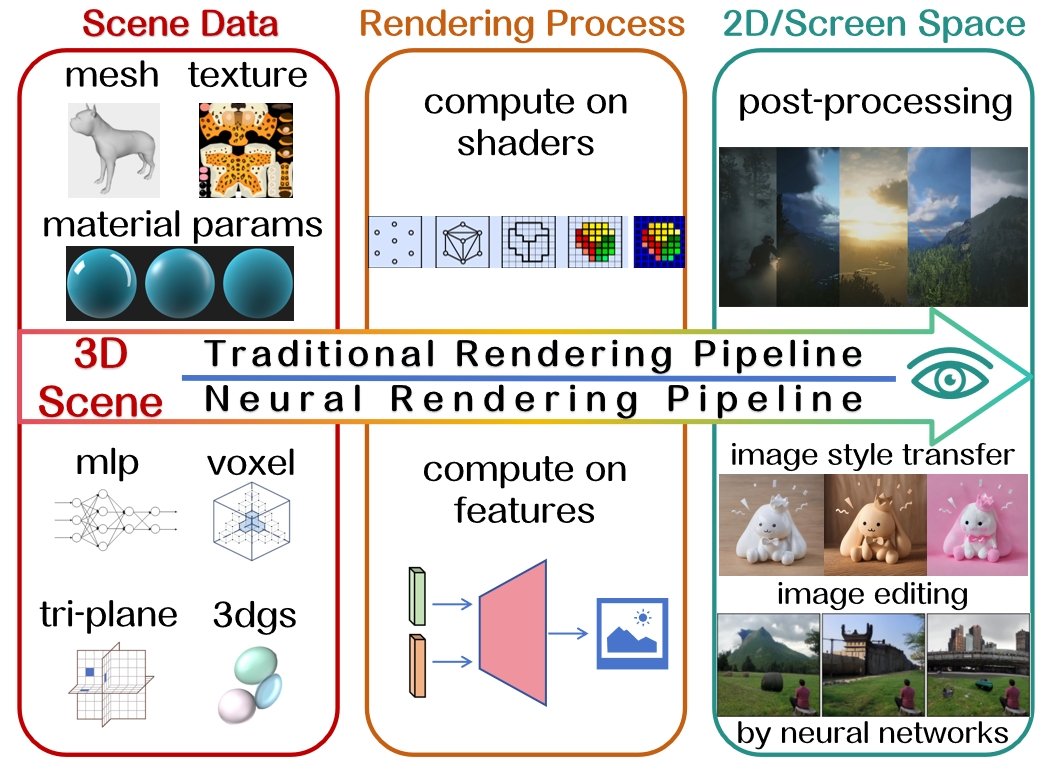}
    \caption{A general overview of mesh-based and radiance field-based rendering pipelines. Images adapted from \cite{yin20213dstylenet,kim2024fprf,chen2024survey,avrahami2022blended,spline_2023}.}
    \label{fig:pipelines}
\end{figure}

This report delves into the latest developments in the creation of 3D digital art using neural stylization methods, as shown in Fig. \ref{fig:teaser}. 
Neural stylization can facilitate automated design and creation of explicit meshes, textures and volumetric assets, which supports seamless usage in the traditional rendering pipeline, accelerating labor-intensive manual tasks such as modeling, texturing, and simulation.
Neural stylization also enables efficient and controllable manipulation or transformation of neural scenes, which typically utilizes neural networks instead of shaders to generate images (Fig.~\ref{fig:pipelines}).
Interestingly, neural stylization has shown practical importance in various applications, including 3D texture design and artistic simulation in movie making \citep{navarro2021stylizing, kanyuk2023singed, hoffman2023creating}, virtual production~\citep{manzaneque2023revolutionizing}, mixed reality experiences \citep{Tseng_2022_SIGGRAPH, taniguchi2019neural}, and artwork creation \citep{varvara2022psy}. 

\input{fig_structure}

% Challenges
Despite its advantages, performing neural stylization in 3D presents new technical challenges such as multi-view consistency, view sampling and rendering, as well as robustness issues including the relative scarcity of 3D datasets, and memory consumption for training and inference.
This report provides a comprehensive discussion and summary of advanced 3D neural stylization techniques that address these challenges, highlighting the power of neural rendering \citep{tewari2022advances}, vision-language models \citep{radford2021learning}, and large-scale generative models \citep{rombach2022high}.

% Structure overview
The structure of this report is depicted in Fig.~\ref{fig_structure}, which is outlined as follows: 
Sec.~\ref{sec:background} reviews fundamentals of 2D neural stylization and important visual or textural feature priors, which act as components or backbones of 3D neural stylization techniques. 
In Sec.~\ref{sec:main}, we introduce a taxonomy for neural stylization and discuss advanced stylization methods on various types of 3D representations in depth with summaries of practical tips to guide future works. 
In Sec.~\ref{sec:eva}, we summarize popular datasets and evaluation metrics for 3D stylization, and particularly, we deliver a benchmark of 3D neural stylization to serve as a reference for the performance of selected methods.
Sec.~\ref{sec:app} introduces diverse applications of 3D neural stylization, demonstrating its practical value in a wide range of domains.
Finally, Sec.~\ref{sec:future} highlights promising research directions with practical significance.

\subsection{Definition and Terminologies}\label{sec:term}
\begin{definition}
\textit{3D neural stylization} refers to the process that employs deep learning techniques and stylization algorithms to generate stylized 3D digital assets or the stylized rendering from these assets, including the alteration of appearance and/or geometry.
\end{definition}
3D neural stylization is well connected to the following terminologies and techniques. 

\noindent$\bullet$~\textbf{Neural Style Transfer (NST)} refers to a class of algorithms that manipulate digital images, or videos, in order to adopt the appearance or visual style of target reference while preserving original content features. It can be regarded as a foundation of 3D neural stylization, as most image-guided 3D stylization methods rely on it. We refer to former surveys \citep{jing2019neural,singh2021neural,zhan2023mise} and provide a concise review in Sec.~\ref{sec:2d-image-guide}.

\noindent$\bullet$~\textbf{Neural Rendering} is a class of techniques that ``learn to render and/or represent a scene from real-world imagery, which can be an unordered set of images, or structured, multi-view images or videos" \citep{tewari2022advances},  several works of which focus on the generation of photorealistic rendering from the neural representations. Instead, neural stylization aims at modifying the visual appearance and aesthetic characteristics of existing digital representations and obtaining artistic or photorealistic rendering results. We refer readers to existing surveys for insight into neural rendering~\citep{tewari2020state,tewari2022advances}, neural fields~\citep{xie2022neural} and image generation~\citep{zhan2023mise}.

\noindent$\bullet$~\textbf{Non-photorealistic Rendering (NPR)} is an area of computer graphics that enables abstract stylized rendering for either 3D models, 3D images or 2D images, such as toon shading, Gooch shading \citep{gooch1998non}, stroke-based painterly rendering \citep{haeberli1990paint, hertzmann1998painterly}, patch-based texture synthesis and transfer \citep{efros2001image, hertzmann2001image}. Mainly leveraging programmable shaders and image-processing techniques, NPR has been widely used in the realms of animation making, digital content creation \citep{MNPR} and game development \citep{NPR_in_game}. However, these techniques require creating handcrafted style patterns and rules, which are labor-intensive and entail domain expertise. In contrast, neural stylization enables fast production with arbitrary style references and has been applied to accelerate cinematic digital production \citep{joshi2017bringing, navarro2021stylizing, hoffman2023creating}.

\noindent$\bullet$~\textbf{Neural Scene Editing} has become more practical in the recent few years thanks to the contribution of large language models (LLMs) and vision-language models (VLMs) \citep{radford2021learning, li2023blip}. Editing methods focus on adding, modifying, or removing objects in a scene, or manipulating some regions of interest. By contrast, stylization methods focus on the transfer of overall appearance and the adoption of specific aesthetic characteristics. Still, stylization methods share critical ideas with editing methods, and some of the methods covered in this survey can also apply to scene editing \citep{koo2023salad,song2023blending, bao2023sine}.

%Previous Surveys and Our Contributions
\subsection{Related Surveys}
In the literature, there exist comprehensive surveys on 2D neural style transfer \citep{jing2019neural, singh2021neural}, surveys on generative image models \citep{zhan2023mise, croitoru2023diffusion, yang2022diffusion}, and surveys on neural field representations~\citep{xie2022neural} and rendering~\citep{tewari2020state, tewari2022advances}. Our survey aims to explore the potential of connecting neural stylization techniques with both traditional and advanced 3D representations, thereby offering valuable resources for style-based 3D digital designs. To the best of our knowledge, this paper is the first comprehensive review to summarize neural stylization techniques and applications specifically tailored to 3D data, highlighting the immense capabilities of neural stylization in the 3D domain.

\section{Background}\label{sec:background}

In this section, we provide a brief discussion on neural style transfer for images, which serves as the fundamental building block for discussing 3D neural stylization (Sec.~\ref{sec:2d-image-guide}). This covers techniques leveraging visual or textual guidance for image style transfer and manipulation, as well as insights on linkages to 3D stylization domain. We also discuss generic methods for 3D content generation with a focus on the state-of-the-art diffusion models for 3D generation (Sec.~\ref{sec:3dgen}).

\begin{figure*}[tbp]
    \centering    
    \includegraphics[width=\linewidth]{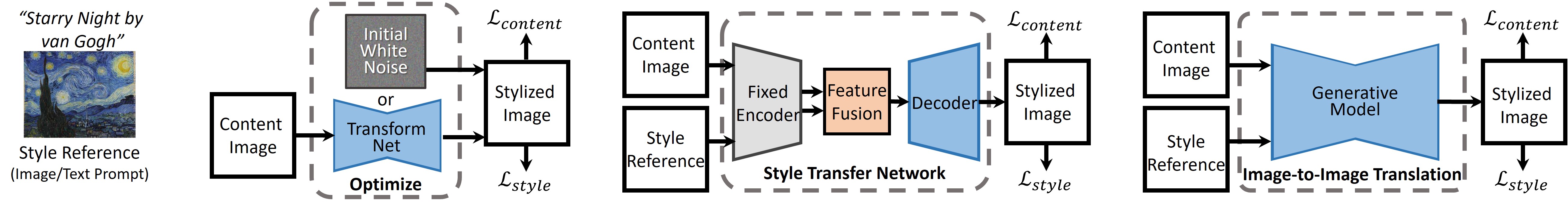}
    \centerline{\footnotesize \quad\quad\quad\quad\quad\quad\quad(a) Single-style optimization \quad\quad\quad\quad\quad (b) Arbitrary style transfer \quad\quad\quad\quad(c) Image-to-image translation\quad}
    \caption{Pipeline comparisons of 2D neural style transfer. (a) Single-style transfer via optimization \citep{gatys2016image, kwon2022clipstyler, johnson2016perceptual}. (b) Arbitrary style transfer via feature fusion or transformation \citep{huang2017arbitrary, li2019learning, liu2021adaattn}. (c) Image-to-image translation with style condition via generative models \citep{huang2018multimodal, deng2022stytr2, wen2023cap, zhang2023inversion}.}
    \label{fig:bg_nst_compare}
\end{figure*}

\subsection{Neural Style Transfer}\label{sec:2d-image-guide}

The basic idea of neural style transfer is to reproduce the style of a reference image for an input image, while keeping the original content of the input. 
The content representation of an image can be extracted by predicting its features using a pre-trained or trainable encoder \citep{simonyan2014very, huang2018multimodal}.
The style representation can be represented by a Gram matrix, which is a dot-product matrix measuring the relevance of each pair of features extracted by a pre-trained network \citep{gatys2016image}. 
Alternatively, the style of an image can also be characterized by the spatially invariant statistics (i.e. channel-wise mean and variance) of features \citep{dumoulin2017learned,huang2017arbitrary}. 
With the rise of vision-language pre-trained models, textual embeddings have been widely employed to represent content or style information \citep{radford2021learning, li2023blip}. In Fig.~\ref{fig:bg_nst_compare}, we provide a high-level pipeline comparison of different types of neural style transfer methods, including single-style transfer via optimization, arbitrary style transfer via feed-forward network, and style transfer via generative models. We discuss each type of method below.

\paragraph{Single Style Transfer}\label{sec:singleST}
One simple method for neural style transfer (Fig.~\ref{fig:bg_nst_compare}a) is to optimize from a white noise image to obtain a new image that shares the content of a source image and the style of a reference image~\citep{gatys2016image}. 
Given a content source image $c$ and a style reference image $s$, the optimization can be done by minimizing a combined objective: \(\mathcal{L}_{total} = \mathcal{L}_{c}(c, cs) + \lambda \mathcal{L}_{s}(s, cs)\),
where the total loss consists of a content loss $\mathcal{L}_{c}$ of the squared-error of VGG features between content image $c$ and output stylized image $cs$, and a Gram matrix style loss $\mathcal{L}_{s}$ between style image $s$ and output stylized image. $\lambda$ is a hyperparameter.

Instead of optimizing an image for each transfer, we can train a single feed-forward network with a perceptual loss to perform style transfer for arbitrary content images~\citep{johnson2016perceptual}. At inference, real-time stylization can be performed simply by forwarding an arbitrary content image through the network. Although performing network inference is much faster than running an optimization, one still needs to retrain the network for each different style.

The rise of vision-language models~\citep{radford2021learning} leads to the possibility of performing style transfer using text prompts as guidance. 
Given CLIP with a text encoder and an image encoder sharing the same latent embedding space~\citep{radford2021learning}, text-guided image style transfer can be achieved by maximizing text-image semantic similarity as the style loss, usually formulated by the CLIP loss defined by
\begin{equation} \label{eq:clip_loss}
    \begin{split}
        \mathcal{L}_{clip}(cs,s_{sty}) &= 1 - sim(\mathit{E_{I}(cs)},\mathit{E_{T}(s_{sty})}),\\
    \end{split}
\end{equation}
where $cs$ and $s_{sty}$ are the stylized image and style text prompt, $E_{I}$ and $E_{T}$ are the pre-trained CLIP image encoder and text encoder, respectively. $sim(A,B)$ is the cosine similarity between two feature vectors.
One can also use the directional CLIP loss~\citep{patashnik2021styleclip, gal2022stylegan} to achieve better style transfer quality:
\begin{equation} \label{eq:directional_clip_loss}
    \begin{split}
        \mathcal{L}_{dir}(c, cs, s_{src}, s_{sty}) &= 1- sim(\Delta I, \Delta T),\\
    \end{split}
\end{equation}
where \(\Delta I = E_{I}(cs) - E_{I}(c), \Delta T = E_{T}(s_{sty}) - E_{T}(s_{src})\), $c$ is the content image, $cs$ is the stylized image, $s_{src}$ and $s_{sty}$ are the source (content) style prompt and target style prompt.
An example of $s_{src}$ and $s_{sty}$ can be ``Photo'' and ``Picasso style painting'', respectively~\citep{kwon2022clipstyler}. 
Since CLIP does not support high-resolution image embedding, a patch-wise version of the directional CLIP loss with augmented patches can be used for better artistic semantic texture transfer~\citep{kwon2022clipstyler}.

\paragraph{Arbitrary Style Transfer: AdaIN and LST} \label{sec:ArbiNST}

To enable the model to transfer arbitrary styles without re-training, one can employ an autoencoder with style fusion (Fig. \ref{fig:bg_nst_compare}b). 
Particularly, to fuse the content and style features, 
we can use adaptive instance normalization (AdaIN) that directly regulates the mean and variance of the feature maps of the content image to match those of the target style image~\citep{huang2017arbitrary}:
    \begin{equation} \label{eq:st_nst_adain}
        AdaIN(c,s) = \sigma(\mathit{F}(s)) \left(\frac{\mathit{F}(c) - \mu(\mathit{F}(c))}{\sigma(\mathit{F}(c))} \right) + \mu(\mathit{F}(s)),
    \end{equation} 
where each VGG feature map $\mathit{F}(\cdot)$ is normalized separately.
The transformed feature maps are fed into a learned decoder to generate the final output.

Alternatively, we can fuse content and style features by learning an affine feature transformation matrix $\mathit{T}$ from content features $\mathit{F}(c)$ and style features $\mathit{F}(s)$ through a convolutional neural network, as proposed by linear style transfer (LST)~
\citep{li2019learning}:
\begin{equation} \label{eq:nst_lst_transform}
    LST(c,s) = \mathit{T} \cdot (\mathit{F}(c) - \mu(\mathit{F}(c))) + \mu(\mathit{F}(s)).
\end{equation}

\paragraph{Generative Models for Style Transfer} \label{sec:2d-text-guide} \label{sec:2d-generative} \label{sec:text_dif}

In addition to optimization-based and feed-forward inference methods, neural style transfer can be achieved via image synthesis tasks such as image-to-image translation and generative models. 
Particularly, image-to-image translation (I2I) achieves style transfer by translating an image from one source domain to a target domain using a generative model such as a generative adversarial network (GAN)~\citep{goodfellow2014generative} (Fig. \ref{fig:bg_nst_compare}c). 
Deterministic I2I translation methods focus on task-specific domain-to-domain translation and do not require style guidance via reference images~\citep{isola2017image, zhu2017unpaired, liu2017unsupervised}. 
Multi-modal I2I translation models enable translation based on examples or latent features \citep{huang2018multimodal, lee2018diverse, chang2020domain, chen2022time}. 
While GAN-based I2I models generate high-fidelity images, they are domain-specific and resource-consuming for training compared to traditional methods (Fig.~\ref{fig:bg_nst_compare}a and \ref{fig:bg_nst_compare}b).

Recently, diffusion models have demonstrated state-of-the-art performance for image synthesis~\citep{rombach2022high}.
A particular strength of diffusion models that contributes to their wide adoption is their ability to learn across different data modalities. 
Similarly to the spirit of CLIP loss for style transfer, text-guided diffusion models leverage textual embedding of the text prompt for conditional image generation and thus allow style transfer via text-to-image generation and text-guided I2I translation~\citep{rombach2022high,saharia2022photorealistic,dalle3_2023}. 
Among the publicly available text-to-image diffusion models, the Stable Diffusion series~\citep{rombach2022high,podell2023sdxl,esser2024scaling} is the most representative. They have inspired a vast amount of research work and a broad range of downstream applications. 

Numerous works explored the potential of text-to-image diffusion models for style transfer. 
One track fine-tunes the diffusion model (usually U-Net) or learns a special textual embedding using a set of images with target style, including techniques like Dreambooth~\citep{ruiz2023dreambooth}, LoRA~\citep{hu2021lora,frenkel2024implicit}, textual inversion~\citep{gal2022image,zhang2023inversion}, etc. 
Among them, B-LoRA~\citep{frenkel2024implicit} jointly fine-tunes two blocks of LoRA layers to capture respectively the style and content of an image, enabling the transfer of this style to unseen content, or such content to new styles.
Textual inversion method InST~\citep{zhang2023inversion} binds a special language mark with a textual embedding inversed from a style image. This style can then be transferred to other images by including the language mark in the prompt during inference.

Another track leverages attention modules in diffusion U-Nets to embed style information without per-style optimization. Spatially invariant feature statistics, as discussed in Sec.~\ref{sec:ArbiNST}, represent style effectively. Diffusion in Style~\citep{everaert2023diffusion} pre-computes the mean and variance for Gaussian noise sampling based on the style feature statistics.
StyleID~\citep{chung2024style} employs AdaIN for noise initialization and replaces content's self-attention \textit{key, value} with those from the style layers. StyleAlign~\citep{hertz2024style} further applies AdaIN to the \textit{query, key} of a sequence of generated images to ensure style consistency. DEADiff~\citep{qi2024deadiff} focuses on cross-attention in high-resolution layers, utilizing Q-Former~\citep{li2023blip} for style extraction instead of AdaIN. InstantStyle~\citep{wang2024instantstyle} isolates styles for transferring by subtracting content CLIP embeddings from their image CLIP embeddings.

Moreover, large pre-trained image editing models such as Instruct-Pix2Pix~\citep{brooks2023instructpix2pix} showcase good style transfer capability. Leveraging LLMs, e.g., GPT-3~\citep{brown2020language}, Instruct-Pix2Pix automates the generation of diverse text prompts and adopts Prompt-to-Prompt~(\citeyear{hertz2022prompt}) to create corresponding image pairs. Subsequently, a standard diffusion model is trained on these pairs, with the exception of concatenating the source image to the first network layer as conditional information. However, its performance degrades when intricate styles are difficult to express in natural language.

\paragraph{Linking 2D to 3D Stylization}\label{sec:link2d3d}

The exploration of 2D neural style transfer offered us valuable insights into style feature and conversion: \textit{spatially invariant statistics (mean and variance) of visual feature maps can represent the image style; one can shift such statistics (to align with those of other images) to control the style of an image}. This insight and several other practical skills can boost 3D stylization research. We briefly exemplify three directions of 3D stylization below.

\begin{enumerate}
    \item \textbf{Loss Function Design.} Sec.~\ref{sec:singleST} revisited the basic version of loss functions for 2D style transfer tasks. When it comes to 3D stylization, it is straightforward to employ similar optimization losses in a view-by-view manner while maintaining multi-view consistency via 3D representation. 
    For instance, \cite{liu2018paparazzi} apply 2D latent content and style losses \citep{gatys2016image} to supervise mesh surface morphing; \cite{chen2024stylecity} optimize texture style jointly with semantics-aware target image features (mean and standard deviation) and textual features (Eq.~\ref{eq:clip_loss}-\ref{eq:directional_clip_loss}).
    \item \textbf{Feed-forward Feature Transform.} 
    The success of feed-forward arbitrary style transfer through feature statistics transformation in the 2D domain inspires feed-forward 3D neural style transfer with 3D-aware feature representation.
    For example, StyleGaussian~\citep{liu2023stylegaussian} applies AdaIN (Eq.~\ref{eq:st_nst_adain}) for VGG~\citep{simonyan2014very} features stored in 3D Gaussians for efficient 3D style transfer. FPRF~\citep{kim2024fprf} applies a semantics-aware local AdaIN for features stored with tri-planes. StyleRF~\citep{liu2023stylerf} proposes a modified volume-adaptive IN for features obtained from feature grids.
    \item \textbf{Stylization with Generative Priors.}
    The burgeoning 2D large generative models (Sec.~\ref{sec:2d-generative}) have been leveraged to handle the stylization and even address 3D consistency issues (with geometry priors).
    A simple yet effective way is to directly stylize a sampled view as the target using pre-trained generative models, followed by backpropagation of the 2D error, such as IN2N~\citep{haque2023instructnerf} and IG2G\citep{igs2gs}. A more advanced approach is score distillation, which leverages the capability of diffusion model to process multi-modal guidance for better controllability. Score distillation was first proposed and widely adopted in 3D generation tasks. We'll discuss the topic in the next part.
\end{enumerate}

\subsection{3D Content Generation} \label{sec:3dgen}

Equipped with a background in neural style transfer, let us now briefly discuss 3D content generation methods, which serve as the background and provide valuable insights for 3D neural stylization subsequently.

\paragraph{3D Representations}
In contrast to image representations, there are various representations for learning to generate 3D content. 
Conventional 3D representations are mostly explicit representations including triangle and polygon meshes, point clouds, and voxel grids (or volumes).  The advances in deep learning have spurred an increasing interest in using neural networks to represent 3D data as neural fields, notably neural radiance fields (NeRF) \citep{mildenhall2020nerf}. Subsequently, there appeared notable hybrid or compact radiance fields representations, featured by neural graphics primitives (NGP) \citep{muller2022instant} and 3D Gaussian splats (3DGS) ~\citep{kerbl20233d}. Some implicit representations, such as signed distance functions (SDF) and their truncated versions (TSDF), also gain popularity to represent implicit shapes. We refer readers to the existing survey for a comprehensive overview of neural fields for visual computing~\citep{xie2022neural}.

\paragraph{3D Generative Models}
Existing 3D generative models have explored different types of 3D representations such as point clouds, voxel grids, meshes, and implicit fields \citep{zhao2021point, qi2017pointnet, wu20153d, masci2015geodesic, chen2019learning}. 3D data-driven generative models are trained with large-scale 3D assets with diverse appearances and shapes, which are challenging to collect~\citep{shapenet2015, deitke2023objaverse, liu2019densepoint}. Inspired by neural volume rendering, there appeared a group of 3D-aware image synthesis works learning 3D generation from accessible 2D data \citep{niemeyer2021giraffe,nguyen2020blockgan,chan2022efficient, gu2022stylenerf}. Since the slow and resource-intensive nature of volume rendering results in long training time and low resolution, one can leverage a reduced 3D representation such as tri-planes in the GAN framework for efficient and high-quality image and 3D data generation~\citep{chan2022efficient}.

\begin{figure}[tbp]
    \centering
    \includegraphics[width=0.99\linewidth]{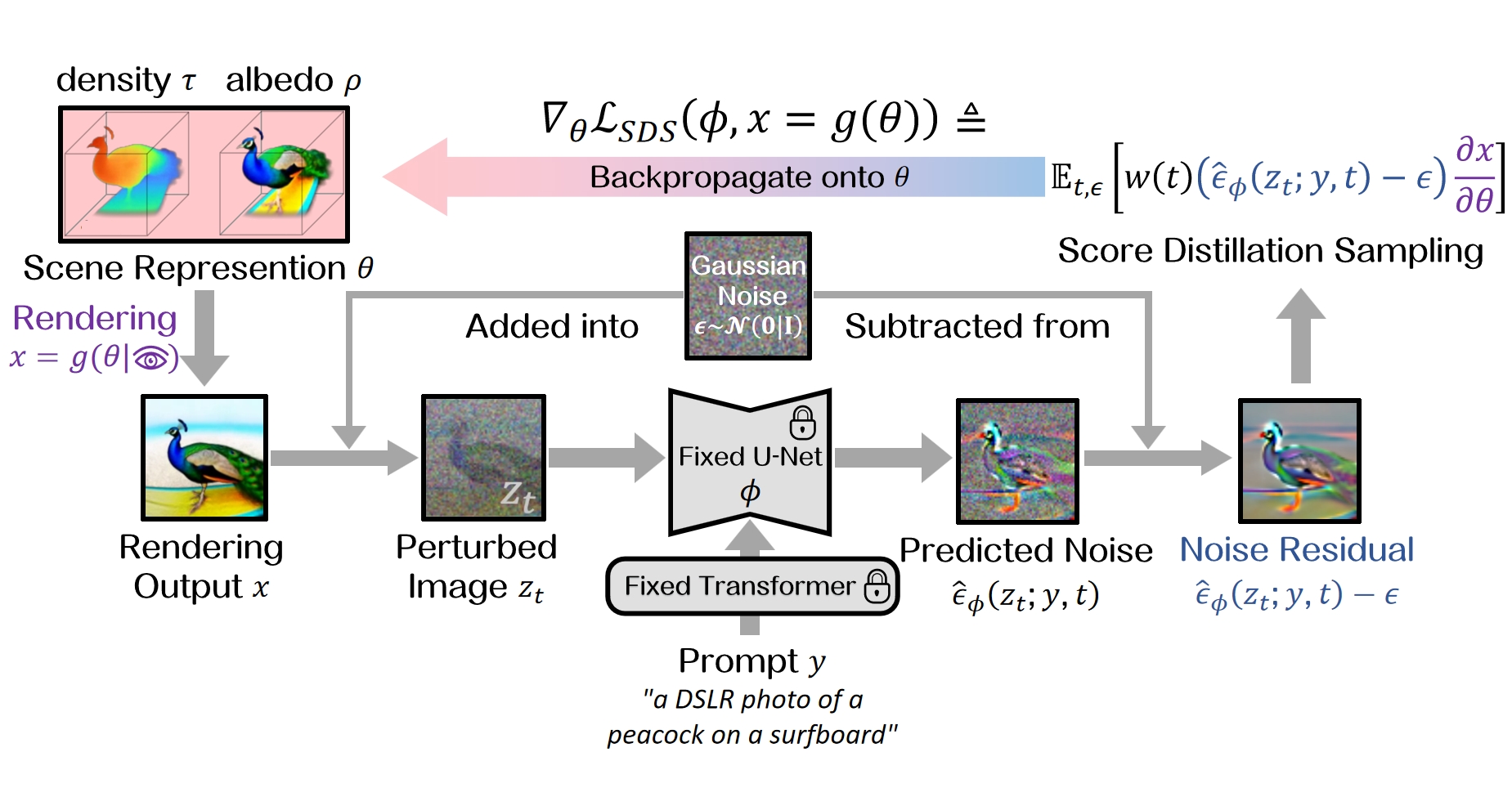}
    \caption{3D generation architecture with score distillation sampling loss. A pre-trained denoising U-Net supervises NeRF optimization. Image adapted from \cite{poole2022dreamfusion}. }
    \label{fig:3dgen}
\end{figure}

\paragraph{3D Generation via Diffusion Priors}

In a similar spirit to 3D-aware image synthesis, it is of great interest to generate 3D data from priors learned by 2D diffusion models~\citep{dalle3_2023,ruiz2023dreambooth}.
DreamFusion~\citep{poole2022dreamfusion} proposed a 3D generation pipeline that optimizes a NeRF by leveraging a text-guided diffusion model as the critic to the NeRF rendered images (Fig.~\ref{fig:3dgen}). The gradient function for optimization is a weighted average of noise residual multiplied by the Jacobian of the rendering process, also known as score distillation sampling (SDS).

Several variants have been proposed to resolve serious problems in the SDS loss such as oversaturation, over-smoothing, and lack of details, generating more realistic and high-definition 3D objects. 
Variational score distillation (VSD) trains a LoRA model to better estimate the data distribution of rendered images for effective updating~\citep{wang2023prolificdreamer}. Delta denoising score (DDS) computes the difference between two SDS scores as guidance~\citep{hertz2023delta}, while posterior distillation sampling (PDS) aligns the stochastic latents of the source image and the target image instead of noise variables~\citep{Koo2024PDS}. Further works explore instilling geometric information to score distillation~\citep{yang20233dstyle, yeh2024texturedreamer}.

Moreover, 3D datasets are also exploited to provide geometric priors such as canonical coordinates map~\citep{li2023sweetdreamer} and normal map~\citep{long2023wonder3d} for better multi-view consistency. 
{Zero-1-to-3}~\citep{liu2023zero} proposed the view-conditioned diffusion that accepts a rotation angle as an extra condition, which synthesizes novel views based on any single view of a 3D model.
Recent 3D generation works further improve the visual quality by combining 2D priors from text-to-image diffusion and 3D-aware priors from view-conditioned diffusion~\citep{qian2023magic123,sun2023dreamcraft3d}.

\begin{figure*}[t]
    \centering
    \includegraphics[width=\linewidth]{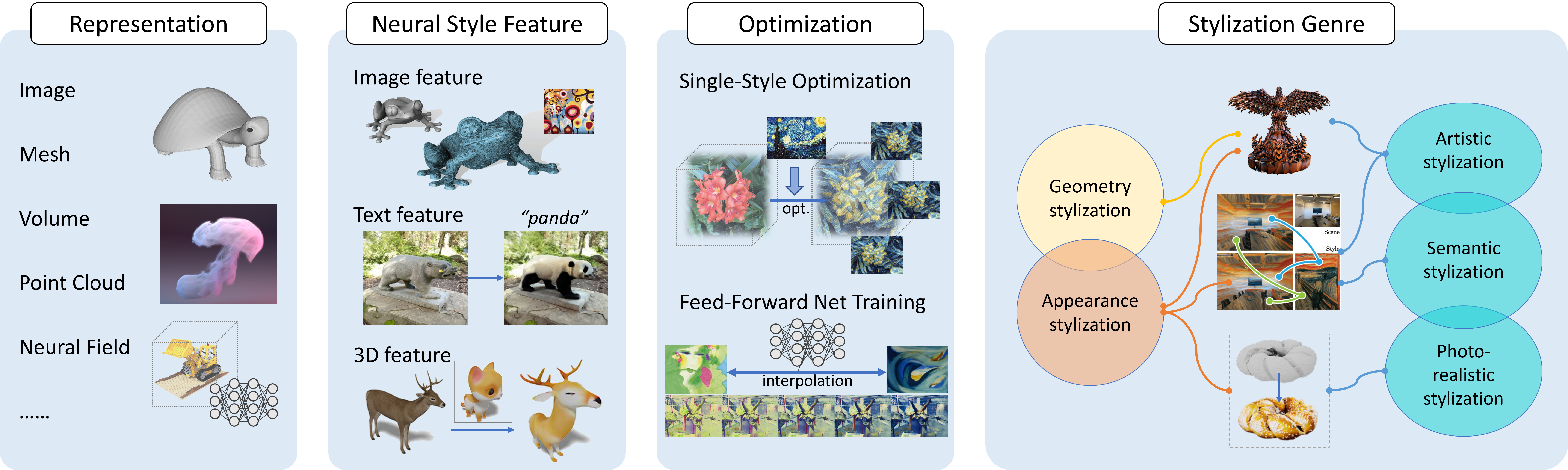}
    \caption{Taxonomy of neural stylization. Images from \cite{richardson2023texture, aurand2022efficient, mildenhall2020nerf, liu2018paparazzi, haque2023instructnerf, yin20213dstylenet, zhang2022arf, liu2023stylerf, chen2023text2tex, pang2023locally}. }
    \label{fig:taxonomy_criteria}
\end{figure*}

\begin{figure*}[t]
\colorlet{linecol}{black!60}
\pgfkeys{/forest,
 rect/.append style   = {rectangle, rounded corners = 2pt,  fill = gray!10}
}
\resizebox{\linewidth}{!}{
\begin{forest}
    for tree={
        line width=1pt,
        align=center,
        child anchor=north,
        parent anchor=south,
        l sep=6mm, % level
        s sep=0.5mm, % sibling
        edge path={
        \noexpand\path[color=linecol, rounded corners=5pt,
          >={Stealth[length=4pt]}, line width=1pt, ->, \forestoption{edge}]
          (!u.parent anchor) --++(0,-4pt) -|
          (.child anchor)\forestoption{edge label};
        },
    } 
 [3D Neural Stylization, rect, 
    [Mesh, rect, name=mesh,
        [Image-guided, rect, name=mesh-image,
            [Single-Style Opt., rect, name=mesh-image-opt,
                [Artistic, rect, name=mesh-image-opt-art,
                    [\makecell{Paparazzi\\(\citeyear{liu2018paparazzi}) \\ StyleMesh\\(\citeyear{hollein2022stylemesh})},  text centered, text width=15mm]],
                [Photorealistic, rect, name=mesh-image-opt-photo,
                    [\makecell{TEXTure\\(\citeyear{richardson2023texture}}),  text centered, text width=20mm]
                ]
            ]
        ]
        [Text-guided, rect, name=mesh-text,
            [Single-Style Opt., rect, name=mesh-text-opt,
                [Artistic, rect, name=mesh-text-opt-art,
                    [\makecell{Text2Mesh(\citeyear{michel2022text2mesh}) \\ X-Mesh(\citeyear{ma2023x})},  text centered, text width=23mm]],
                [Photorealistic, rect, name=mesh-text-opt-photo,
                    [\makecell{TEXTure(\citeyear{richardson2023texture})\\ Text2Tex(\citeyear{chen2023text2tex}) \\ Text2Scene(\citeyear{hwang2023text2scene})\\TextDeformer(\citeyear{gao2023textdeformer}) \\ Text2Mesh(\citeyear{michel2022text2mesh})\\X-Mesh(\citeyear{ma2023x})},  text centered, text width=30mm]]
            ]
        ]
    ]
    [Neural Field, rect, name=field,
        [Image-guided, rect, name=field-image,
            [Single-Style Opt., rect, name=field-image-opt,
                [Artistic, rect, name=field-image-opt-art,
                    [\makecell{ARF(\citeyear{zhang2022arf})\\ INS(\citeyear{fan2022unified}) \\ SNeRF(\citeyear{nguyen2022snerf})\\ Ref-NPR(\citeyear{zhang2023ref})\\ LocalStyleNeRF\\(\citeyear{pang2023locally})},  text centered, text width=23mm]],
                [Photorealistic, rect, name=field-image-opt-photo,
                    [\makecell{LipRF(\citeyear{zhang2023transforming})\\ SNeRF(\citeyear{nguyen2022snerf})},  text centered, text width=20mm]
                ]
            ]
            [Feed-forward Net Training, rect, name=field-image-train,
                [Artistic, rect, name=field-image-train-art,
                    [\makecell{LSNV(\citeyear{huang2021learning})\\ HyperStyle(\citeyear{chiang2022stylizing})\\ StylizedNeRF(\citeyear{huang2022stylizednerf}) \\StyleRF(\citeyear{liu2023stylerf})},  text centered, text width=27mm]],
                [Photorealistic, rect, name=field-image-train-photo
                    [\makecell{UPST\\(\citeyear{chen2022upst})},  text centered, text width=13mm]
                ]
            ]
        ]
        [Text-guided, rect, name=field-text,
            [Single-Style Opt., rect, name=field-text-opt,
                [Artistic, rect, name=field-text-opt-art,
                    [\makecell{TSNeRF(\citeyear{wang2023tsnerf})\\  NeRF-Art(\citeyear{wang2023nerf}) \\ Instruct-N2N\\(\citeyear{haque2023instructnerf})\\ BlendingNeRF\\(\citeyear{song2023blending})},  text centered, text width=25mm]],
                [Photorealistic, rect, name=field-text-opt-photo,
                    [\makecell{NeRF-Art\\(\citeyear{wang2023nerf}) \\ Instruct-N2N\\(\citeyear{haque2023instructnerf}) \\ BlendingNeRF\\(\citeyear{song2023blending})},  text centered, text width=21mm]
                ]
            ]
        ]
    ]
    [Volume, rect, name=vol,
        [Image-guided, rect, name=vol-image,
            [Single-Style Opt., rect, name=vol-image-opt,
                [Artistic, rect, name=vol-image-opt-art,
                    [\makecell{TNST(\citeyear{kim2019transport}) \\ LNST(\citeyear{kim2020lagrangian})},  text centered, text width=18mm]
                ]
            ]
            [Feed-forward Net Training, rect, name=vol-image-train,
                [Artistic, rect, name=vol-image-train-art,
                    [\makecell{SKPN(\citeyear{guo2021volumetric}) \\ENST(\citeyear{aurand2022efficient})},  text centered, text width=20mm]
                ]
            ]
        ]
    ]
  ]
\end{forest}
}
\caption{Hierarchical classification of selected image- and text-guided 3D neural stylization methods.} \label{fig:taxonomy_methods} 
\end{figure*}

\section{3D Neural Stylization}\label{sec:main}
% classification: 1.3d representation type;  2.guidance type  3.stylize 3D assets or just get stylized rendering results without changing original content  4. applicable scenarios  5.online/offline
% give a table for every subsection rather than a whole

In this section, we first establish a taxonomy for neural stylization and give an example of the categorization of selected 3D neural stylization methods (Sec.~\ref{sec:3d_style_tax}). 
In the subsequent sections, we will discuss state-of-the-art 3D neural stylization techniques on diverse 3D representations, such as meshes (Sec.~\ref{sec:3d_style_mesh}), neural fields (Sec.~\ref{sec:3d_style_neural_field}), volumetric data (Sec.~\ref{sec:3d_style_volume}), point clouds (Sec.~\ref{sec:3d_style_point}), and implicit shapes (Sec.~\ref{sec:3d_style_implicit_shape}). We then discuss a set of guidelines for practical implementations of 3D stylization (Sec.~\ref{sec:practical}). 

\subsection{Taxonomy} \label{sec:3d_style_tax}

Our taxonomy for neural stylization methods consists of the following aspects: 
\begin{itemize}
\item \textit{Representations}. We categorize stylization methods based on data representations such as image, mesh, volume, point cloud, and neural field. 
\item \textit{Neural Style Feature}. We categorize based on image visual features, textual semantic features, or 3D latent features derived from pre-trained models, typically neural classifiers or generative models.
\item \textit{Optimization}. This refers to optimization-based or prediction-based stylization methods with single, multiple, or arbitrary styles supported.
\item \textit{Stylization Genres}. This refers to different types of stylization, mainly including geometry stylization operating on asset shape and surface patterns, and appearance stylization focusing on color, texture and visual patterns to align with specific styles from artistic paintings to realistic concepts.

\end{itemize}
To guide the reader through the main section of this survey, we illustrate the taxonomy in Fig.~\ref{fig:taxonomy_criteria}, and provide a hierarchical classification of the 3D stylization methods in Fig.~\ref{fig:taxonomy_methods}. 
Let us now discuss 3D neural stylization methods by following the categorization based on 3D representations below.

\begin{table*}[tbp]
    \setlength{\tabcolsep}{1pt}
    \centering
    \resizebox{\linewidth}{!}{ 
    \begin{tabular}{l ccccccc}
         \toprule
         Method &  Input & Guidance& Output & Scene & Optimization  & Style Genre & \faCode \\           
         \midrule
         3DStyleNet~(\citeyear{yin20213dstylenet})& textured mesh & \makecell{points, image}& textured mesh & object & feed-forward & \makecell{geo.$\&$app.} & \xmark \\
         Point-UV Diffusion(\citeyear{yu2023texture})& mesh & points, image or text& textured mesh& object & feed-forward & appearance & \href{https://github.com/CVMI-Lab/Point-UV-Diffusion}{\faExternalLink} \\ 
         TexDreamer~(\citeyear{liu2024texdreamer})& human mesh & image or text& texture & human &feed-forward&  appearance &  \href{https://github.com/ggxxii/texdreamer}{\faExternalLink}\\
         
         Paparazzi~(\citeyear{liu2018paparazzi})& mesh & image & mesh & object & single-style & geometry, artistic & \href{https://github.com/HTDerekLiu/Paparazzi}{\faExternalLink} \\ 
         StyleMesh~(\citeyear{hollein2022stylemesh})& RGBD imgs, rec. mesh & image & texture & \makecell{room} &  single-style & appearance, artistic & \href{https://github.com/lukasHoel/stylemesh}{\faExternalLink}\\ 
         
         Text2Mesh~(\citeyear{michel2022text2mesh})& mesh & text or image & vertex-colored mesh & object & single-style & \makecell{geo.$\&$app.} &\href{https://github.com/threedle/text2mesh}{\faExternalLink}\\ 
         X-Mesh~(\citeyear{ma2023x})& mesh & text & vertex-colored mesh & object  & single-style&  \makecell{geo.$\&$app.}&\href{https://github.com/xmu-xiaoma666/X-Mesh}{\faExternalLink}\\ % ICCV2023 
        
         Text2Scene~(\citeyear{hwang2023text2scene})& meshes, segment labels & \makecell{image \& text} & vertex-colored mesh & room w/ objects  & single-style & appearance & \href{https://github.com/uvavision/Text2Scene}{\faExternalLink}\\ % CVPR 2023 

         TextDeformer(\citeyear{gao2023textdeformer})& mesh & text & mesh & object & single-style & geometry & \href{https://github.com/threedle/TextDeformer}{\faExternalLink}\\ % SG 2023 Conf
        
         TEXTure~(\citeyear{richardson2023texture})& mesh & text or image & texture & object & single-style & appearance & \href{https://github.com/TEXTurePaper/TEXTurePaper}{\faExternalLink}\\ % SGConf 2023  
         Text2Tex~(\citeyear{chen2023text2tex})& mesh & text & texture & object & single-style& appearance & \href{https://github.com/daveredrum/Text2Tex}{\faExternalLink}\\ % ICCV 2023 
         TexFusion~(\citeyear{cao2023texfusion})& mesh & text & texture & object & single-style & appearance & \xmark\\ % ICCV 2023 
         3DStyle-Diffusion(\citeyear{yang20233dstyle})& mesh& text& neural texture & object& single-style& appearance & \href{https://github.com/yanghb22-fdu/3dstyle-diffusion-official}{\faExternalLink} \\
         SceneTex~(\citeyear{chen2023scenetex}) & mesh & text & texture & room & single-style & appearance & \href{https://github.com/daveredrum/SceneTex}{\faExternalLink}\\
         Paint-it~(\citeyear{youwang2023paint})& mesh& text& PBR texture & object & single-style& appearance& \href{https://github.com/postech-ami/paint-it}{\faExternalLink}\\
         Paint3D~(\citeyear{zeng2023paint3d})& mesh& image or text & texture & object &single-style &appearance & \href{https://github.com/OpenTexture/Paint3D}{\faExternalLink}\\
         TeMO~(\citeyear{zhang2024temo})& multi-meshes& text & textures & objects&single/multi-style &appearance & \href{https://github.com/zhangxuying1004/TeMO}{\faExternalLink}\\
         EASI-Tex~(\citeyear{perla2024easitex})& mesh & image & texture&object &single-style & appearance& \href{https://github.com/sairajk/easi-tex}{\faExternalLink}\\
         DreamMat~(\citeyear{zhang2024dreammat})& mesh & text & PBR texture& object&single-style & appearance& \href{https://github.com/zzzyuqing/DreamMat}{\faExternalLink} \\
         FlashTex~(\citeyear{deng2024flashtex})& mesh &  text & PBR texture & object &single-style&  appearance &  \href{https://github.com/Roblox/FlashTex}{\faExternalLink}\\
         StyleCity~(\citeyear{chen2024stylecity})& textured mesh& image $\&$ text& texture& city & single-style& appearance &  \href{https://github.com/chenyingshu/stylecity3d}{\faExternalLink}\\
         ControllableNST~(\citeyear{gomes2024controllable})& static/dynamic mesh& image &static/dynamic (colored) mesh &object(s) &single-style & geo.(\&app.)& \xmark\\         
         % & & & & & & & \\
         \bottomrule
    \end{tabular}}
    \caption{Summary of selected mesh-based neural stylization. PBR refers to physically-based rendering.} 
    \label{tab:sec_mesh_style}
\end{table*}
\subsection{Mesh-based Stylization} \label{sec:3d_style_mesh}

In computer graphics and 3D modeling, a mesh is a collection of vertices, edges, and faces that define the geometric structure of an object. Objects represented by meshes can also store additional appearance attributes, such as vertex colors, materials, UV coordinates, and texture maps. Additionally, neural networks, such as multi-layer perceptrons (MLPs), can represent these attributes, including neural textures~\citep{thies2019deferred, oechsle2019texture}, neural reflectance field \citep{baatz2022nerf}, neural visibility field \citep{srinivasan2021nerv}, neural vertex \citep{michel2022text2mesh, ma2023x, lei2022tango}, etc. By using differentiable renderers \citep{ravi2020pytorch3d, Laine2020diffrast, KaolinLibrary}, we can optimize these explicit or implicit attribute representations for 3D geometry manipulation and appearance editing. For example, one can predict vertex positions and colors \citep{michel2022text2mesh}, SVBRDF parameters and normals \citep{lei2022tango}, and synthesize new texture images \citep{richardson2023texture}.
The following sections cover critical techniques for mesh-based stylization, including geometric deformation~(Sec.~\ref{sec:3d_style_geo_mesh}) and texture synthesis~(Sec.~\ref{sec:3d_style_tex}) to align with a provided image, text, or 3D shape guidance.
Table \ref{tab:sec_mesh_style} shows a comparison of recent mesh-based stylization methods.

\subsubsection{Surface Geometric Deformation} \label{sec:3d_style_geo_mesh}
3D neural stylization enables deforming mesh geometry to align with artistic visual patterns or a specified shape, guided by visual references or textual descriptions. This capability facilitates creative 3D modeling such as surface engraving effect~\citep{liu2018paparazzi} and geometry morphing \citep{gao2023textdeformer}.  
Existing works learn geometry variations in the form of vertex position displacement, e.g., explicit displacement via differentiable rendering \citep{liu2018paparazzi} and implicit displacement via neural networks \citep{michel2022text2mesh, ma2023x, gao2023textdeformer}. 

For example, {Paparazzi} is a neural stylization method based on differentiable rendering that allows the propagation of changes in the image domain to changes of the mesh vertex positions~\citep{liu2018paparazzi}. It takes a triangle mesh as input, and applies latent VGG content and style losses (Sec. \ref{sec:2d-image-guide}, \cite{gatys2016image}) between rendered image(s) and gray-scale style image to update vertex positions. After convergence, the mesh surface is stylized with artistic strokes and motifs from the style image.

With CLIP loss (Sec.~\ref{sec:2d-text-guide}, Eq.~\ref{eq:clip_loss}), recent works explored text-guided mesh geometric and/or appearance alteration \citep{michel2022text2mesh, ma2023x}.
{Text2Mesh} (\citeyear{michel2022text2mesh}) and {X-Mesh} (\citeyear{ma2023x}) incorporate Neural Style Field, which is composed of an MLP network that maps vertex coordinates to vertex color (offset) and vertex position offset. The stylized mesh, with updated vertex colors and vertex positions, is rendered into multiple colored and gray-scale images, which are used for computing CLIP loss against a given text prompt. 

Besides updating with the CLIP loss, {X-Mesh} \citep{ma2023x} employs an attention module to directly include the prompt CLIP embedding as additional input along with the vertex coordinate embeddings. 
Accordingly, {X-Mesh} achieves fast convergence in a few minutes for high-quality stylized results. 
{TextDeformer}~\citep{gao2023textdeformer} upgraded the local CLIP-guided mesh geometric stylization \citep{wang2022clip, michel2022text2mesh} through Jacobians for global and smooth mesh deformation  \citep{aigerman2022neuraljacobian}. 
Instead of learning position displacement directly, they assign Jacobians by matrices for each triangle and solve a Poisson problem \citep{aigerman2022neuraljacobian} to compute the corresponding vertex deformation map, which largely achieves deformation with low-frequency to high-frequency details.

Alternatively, {3DStyleNet}~\citep{yin20213dstylenet} learns to perform joint geometric and texture style transformation from one 3D object to another, and interpolation of geometric and texture style. This method consists of a 3D geometric part-aware style transfer network and a 2D texture style transfer network. The authors innovatively abstracted the geometries of an object with a set of 3D Gaussian ellipsoids and employed a learned 3D part-aware semantics affine transformation field based on Linear Blend Skinning (LBS) model \citep{lindholm2001user}. Meanwhile, they transfer the mesh texture using regular image style transfer techniques (Sec. \ref{sec:2d-image-guide} LST (\citeyear{li2019learning})). The two networks are pre-trained with non-textured mesh models \citep{turbosquid_2023, shapenet2015, renderpeople_2023} and images (WikiArt (\citeyear{kaggle_2016}) and COCO (\citeyear{lin2014microsoft})) respectively. They are then jointly optimized by part-aware, content and style losses through rendered multi-views via a differentiable renderer \citep{Laine2020diffrast}. 

A recent work~\citep{gomes2024controllable} further explores geometry stylization for dynamic meshes, enabling efficient production of physics simulation and animation. They employ neural neighbor style transfer~\citep{kolkin2022neural} instead of Gram-matrix to guide the style transfer, which produces higher quality high-frequency details by replacing each individual feature of the content image with its closest feature of the style image. The key to an effective and natural global and local stylization is the multi-level parameterization of mesh vertex position, which allows the 2D error to propagate in differential rendering sufficiently. An additional mechanism for vertex displacement interpolation and smoothing across frames is applied to improve time coherency.  These enhancements result in high-quality, artifact-free mesh stylizations, suitable for creating unique artistic looks in simulations and 3D asset design.

\subsubsection{Texture Synthesis}\label{sec:3d_style_tex}
Mesh textures are essential for representing complex visual appearance with color and patterns, for which image- or text-guided neural style transfer are well developed (Sec.~\ref{sec:2d-image-guide}). Several methods have explored texture transformation and synthesis using visual features such as VGG features and diffusion priors \citep{hollein2022stylemesh, lei2022tango, richardson2023texture, cao2023texfusion, yang20233dstyle}.
We categorize these methods based on their optimization and learning techniques and discuss them below.

\begin{figure}[tbp]
    \centering    
    \begin{subfigure}[b]{0.45\linewidth}
         \centering
         \includegraphics[width=\linewidth]{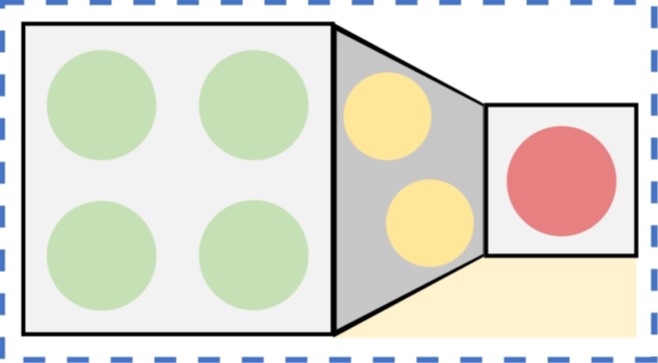}
         \caption{\footnotesize Projected geometry}
     \end{subfigure}
     \hfill
    \begin{subfigure}[b]{0.45\linewidth}
         \centering
         \includegraphics[width=\linewidth]{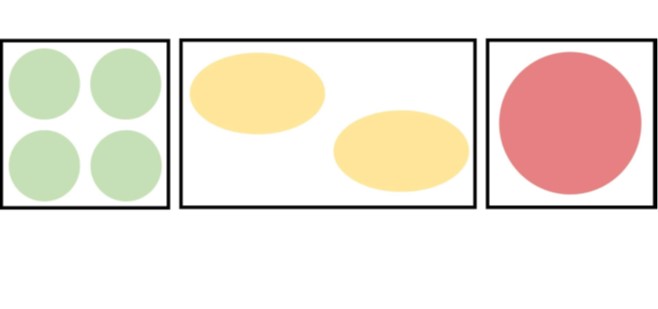}
         \caption{\footnotesize World-space geometry}
     \end{subfigure}
    \caption{Stretched pattern artifacts from stylization in screen space. Image adapted from \cite{kato2018neural}.}
    \label{fig:stylemesh_stretch_ex}
\end{figure}

\noindent\textbf{Optimize via 2D Features.} With an artistic image reference, StyleMesh~\citep{hollein2022stylemesh} proposed a depth- and angle-aware texture optimization scheme for the 3D reconstructed indoor room. It optimizes an explicit texture image by backpropagating gradients computed from 2D content and style losses (Sec. \ref{sec:2d-image-guide}, \cite{gatys2016image}) between each view of the scene and the style reference image. 
This method leverages depth and normal information from the mesh, mitigating artifacts such as view-dependent stretch and size artifacts that commonly arise from conventional 2D losses in 3D scenarios, as shown in Fig.~\ref{fig:stylemesh_stretch_ex}.  
Nonetheless, {StyleMesh} largely relies on posed images under reconstructed scenes and ground-truth depths.

Optimizing mesh appearance colors with style descriptions using CLIP is effective but may not always achieve realistic results~\citep{michel2022text2mesh, lei2022tango, ma2023x}. 
Recently, text-to-image diffusion models have gained popularity for their ability to synthesize high-fidelity images. Therefore, researchers start to explore lifting 2D diffusion priors for 3D generation \citep{poole2022dreamfusion, wang2023prolificdreamer, lin2023magic3d, chen2023fantasia3D} and stylization \citep{chen2023text2tex, yang20233dstyle, zeng2023paint3d, youwang2023paint}. 
Among these, TEXTure~\citep{richardson2023texture} is a text-guided 3D texture painting method that iteratively paints the texture in a view-by-view manner. To maintain 3D consistency, each view drawing iteration is guided by a view-dependent trimap that indicates the ``keep", ``refine", and ``generate" regions to control the amount of newly generated content for the texture. 
Alongside the trimap, a rendered RGB and depth map are fed into a pre-trained depth-to-image diffusion model (Sec.~\ref{sec:text_dif}, ControlNet (\citeyear{zhang2023adding})) to obtain a synthesized view. This synthesis is finally projected back to the texture map via optimization. Apart from texturing, {TEXTure} supports various tasks such as texture transfer, texture editing, and multi-view image transfer.

A concurrent work to TEXTure is {Text2Tex}~\citep{chen2023text2tex}, which inpaints likewise the texture images from different views progressively with the help of the confident trimap and a depth-to-image ControlNet 
 (\citeyear{zhang2023adding}). 
This method presets some axis-aligned viewpoints and alternatively updates the next best view (see Table \ref{tab:3d_disc_view_plan}), which follows a more robust automatic view scheduling strategy addressing the blurriness and stretching artifacts.
Paint3D~\citep{zeng2023paint3d} employs coarse-to-fine UV diffusion models to further refine incomplete areas of multi-view inpainted texture in high definition.
Based on an indoor room scenario, DreamSpace \citep{yang2023dreamspace} synthesizes an indoor panorama with additional inpainting with diffusion models \citep{zhang2023adding} for more consistent texture synthesis.

% Method2:  optimize latent feature by diffusion synthesis
\noindent\textbf{Optimize in Latent Space.} While running the entire generative diffusion process for multi-view painting provides an efficient approach for texture synthesis, it often results in inconsistent texture patterns and overall style. Instead, {TexFusion} \citep{cao2023texfusion} updates a 3D consistent latent texture at each denoising step from multi-views conditioned on previous denoising steps. To ensure consistency, the final texture image is optimized by distilling multi-view images decoded by a pre-trained depth-conditioned diffusion model (Sec.~\ref{sec:text_dif}, Stable Diffusion (\citeyear{rombach2022high})) and with a neural color field mapping 3D coordinates to RGB values \citep{muller2022instant}. Similarly, \citet{knodt2023consistent} updates a latent texture map from multi-view via MultiDiffusion \citep{bar2023multidiffusion}, a multi-window joint diffusion technique for multi-view consistency.

% Method3:  score distillation
\noindent\textbf{Optimize via Score Distillation.} Inspired by score distillation sampling (SDS) techniques (Sec.~\ref{sec:3dgen}, Fig. \ref{fig:3dgen}), several works employ SDS and its variants for texture optimization, focusing on a single object or a single room~\citep{yang20233dstyle, guo2023decorate3d, wu2023hyperdreamer, chen2023scenetex, yeh2024texturedreamer}. In particular, TextureDreamer~\citep{yeh2024texturedreamer} and 3DStyle-Diffusion~\citep{yang20233dstyle} adopt a neural field representation with BRDF parameters \citep{lei2022tango} to facilitate photorealistic rendering. Both works \citep{yang20233dstyle, yeh2024texturedreamer} incorporate geometry-conditioned score distillation from ControlNet (\citeyear{zhang2023adding}), leveraging additional inputs such as depth, normal, and camera pose. Additionally, Decorate3D~\citep{guo2023decorate3d} and HyperDreamer~\citep{wu2023hyperdreamer} utilize super-resolution diffusion techniques to enhance the synthesis of textures at higher resolutions.

\noindent\textbf{Optimize with 3D Shape Supervision.} {Point-UV Diffusion}~\citep{yu2023texture} explores texture synthesis leveraging the shape attributes such as vertex coordinates, normals, and segment masks of the mesh model. The proposed coarse-to-fine texture synthesis framework that combines a point diffusion network \citep{liu2019point, zhou20213d} and a UV diffusion network enables unconditioned texture synthesis for arbitrary mesh models of each training category in the ShapeNet dataset \citep{shapenet2015}. The pipeline can also receive additional image or text guidance using the CLIP encoder. Given a mesh and style visual or textual guidance, the point diffusion model generates color for sampled points from the mesh, which are then projected onto 2D UV space to create a coarse texture image. Subsequently, the UV diffusion model utilizes the coarse texture and additional shape attributes to predict the high-fidelity texture. 

\begin{table*}[tbp]
    \centering
    \resizebox{\linewidth}{!}{ 
    \begin{tabular}{l l ccccc c}
         \toprule
         Method & Base Technique  & 3D Repr.(Struct.) & Guidance & Scene & Optimization  & Style Genre & \faCode \\  
         
         \midrule
         LSNV~(\citeyear{huang2021learning}) &FVS(\citeyear{riegler2020free}), LST(\citeyear{li2019learning})  & point cloud & image & bounded/unbounded &  feed-forward & appearance, artistic & \href{https://github.com/hhsinping/stylescene}{\faExternalLink}\\ 
         HyperStyle (\citeyear{chiang2022stylizing}) &NeRF++(\citeyear{zhang2020nerf++}), HyperNet(\citeyear{ha2016hypernetworks}) &NeRF & image & unbounded & feed-forward & appearance, artistic  & \href{https://github.com/ztex08010518/Stylizing-3D-Scene}{\faExternalLink}\\ 
         StyleRF~(\citeyear{liu2023stylerf}) & TensoRF(\citeyear{chen2022tensorf}), SICT(\citeyear{liu2023stylerf}), LST(\citeyear{li2019learning}) &NeRF (grid) & images & bounded & feed-forward & appearance, artistic  & \href{https://github.com/Kunhao-Liu/StyleRF}{\faExternalLink}\\ % CVPR 2023
         StylizedNeRF(\citeyear{huang2022stylizednerf})& NeRF(\citeyear{mildenhall2020nerf}), AdaIN(\citeyear{huang2017arbitrary})  &NeRF & image & bounded/unbounded &  feed-forward & appearance, artistic  & \href{https://github.com/IGLICT/StylizedNeRF}{\faExternalLink}\\ % 

         UPST-NeRF (\citeyear{chen2022upst}) & DVGO(\citeyear{sun2022direct}), I2I-GAN  &NeRF (grid) & image  & bounded & feed-forward & appearance, photoreal  & \href{https://github.com/semchan/UPST-NeRF}{\faExternalLink}\\ 
         FPRF (\citeyear{kim2024fprf}) & K-Planes(\citeyear{fridovich2023k}),AdaIN(\citeyear{huang2017arbitrary}) & NeRF (grid) & images & unbounded & feed-forward & appearance, photoreal  & \href{https://kim-geonu.github.io/FPRF}{\faExternalLink}\\ % AAAI2024 
        StyleGaussian (\citeyear{liu2023stylegaussian}) & 3DGS(\citeyear{kerbl20233d}),AdaIN(\citeyear{huang2017arbitrary}) & 3DGS & images &bounded/unbounded  & feed-forward &appearance &\href{https://github.com/Kunhao-Liu/StyleGaussian}{\faExternalLink} \\ %SA2024 TC
         
         % ARF, INS (ECCV22)
         ARF (\citeyear{zhang2022arf}) & Plenoxels(\citeyear{yu_and_fridovichkeil2021plenoxels}), NNFM(\citeyear{zhang2022arf})  &NeRF (grid) & image  & bounded/unbounded & single-style & appearance, artistic  & \href{https://github.com/Kai-46/ARF-svox2}{\faExternalLink}\\
         INS \citeyear{fan2022unified} & NeRF(\citeyear{mildenhall2020nerf}), Gatys \etal(\citeyear{gatys2016image})  &NeRF, SDF & image & bounded  &  single/multi-style & appearance, artistic  & \href{https://github.com/VITA-Group/INS}{\faExternalLink}\\ % 
         
         SNeRF (\citeyear{nguyen2022snerf}) & any NeRF, any NST &NeRF & image  & bounded/unbounded  & single-style & app.($\&$geo.), artistic  & \xmark\\ %SG22 
         % LipRF, StyleRF, Ref-NPR (CVPR23)
         LipNeRF (\citeyear{zhang2023transforming}) & Plenoxels(\citeyear{yu_and_fridovichkeil2021plenoxels}) &NeRF (grid) & image & bounded/unbounded &  single-style & appearance, photoreal  & \xmark\\ 
         Ref-NPR(\citeyear{zhang2023ref}) & Plenoxels(\citeyear{yu_and_fridovichkeil2021plenoxels}), any 2D edit  &NeRF (grid) & image(s) & bounded/unbounded & single-style & appearance  & \href{https://github.com/dvlab-research/Ref-NPR/}{\faExternalLink}\\ %CVPR 2023
         LocalStyleNeRF (\citeyear{pang2023locally})& iNGP(\citeyear{muller2022instant}), NNFM(\citeyear{zhang2022arf}) & NeRF(hash grid) & image & room & single/multi-style & appearance, artistic  & \href{https://github.com/hkust-vgd/nerfstyle}{\faExternalLink}\\ % ICCV23 
         
         SINE (\citeyear{bao2023sine}) &NeRF(\citeyear{mildenhall2020nerf}),DIF(\citeyear{deng2021deformed}), DINO(\citeyear{caron2021emerging}) &NeRF  & image  & bounded/unbounded & single-style & geo.$\&$app. & \href{https://github.com/zju3dv/SINE}{\faExternalLink}\\ % CVPR2023        
         NeRF-Art (\citeyear{wang2023nerf}) & VolSDF(\citeyear{yariv2021volume}), CLIP(\citeyear{radford2021learning})  &NeRF & text & bounded & single-style & appearance  & \href{https://github.com/cassiePython/NeRF-Art}{\faExternalLink}\\ % TVCG23
         Instruct-N2N (\citeyear{haque2023instructnerf}) & Nerfacto(\citeyear{tancik2023nerfstudio}), InstructPix2Pix(\citeyear{brooks2023instructpix2pix})  & NeRF (hash grid) & text & bounded/unbounded & single-style & geo.$\&$app. &\href{https://github.com/ayaanzhaque/instruct-nerf2nerf}{\faExternalLink}\\ % ICCV23 
         BlendingNeRF (\citeyear{song2023blending}) &NeRF(\citeyear{mildenhall2020nerf}), CLIP(\citeyear{radford2021learning}) &NeRF & text & object & single-style & geo.$\&$app.  & \xmark \\ % ICCV23 
         Vica-NeRF (\citeyear{dong2023vica})& Nerfacto(\citeyear{tancik2023nerfstudio}), InstructPix2Pix(\citeyear{brooks2023instructpix2pix})  & NeRF (hash grid) & text & bounded/unbounded & single-style & geo.$\&$app.  & \href{https://github.com/Dongjiahua/VICA-NeRF}{\faExternalLink}\\
         GaussianEditor (\citeyear{chen2023gaussianeditor})& 3DGS(\citeyear{kerbl20233d}), any edit& 3DGS & text &bounded/unbounded & single-style & geo.$\&$app. & \href{https://github.com/buaacyw/GaussianEditor}{\faExternalLink} \\ %CVPR24
         GaussCtrl (\citeyear{gaussctrl2024})& 3DGS(\citeyear{kerbl20233d}), any edit& 3DGS & text &bounded/unbounded & single-style & geo.$\&$app. & \href{https://github.com/ActiveVisionLab/gaussctrl}{\faExternalLink}\\
         GaussianGrouping (\citeyear{ye2025gaussian})& 3DGS(\citeyear{kerbl20233d}), any edit& 3DGS & text &bounded/unbounded & single-style & geo.$\&$app. & \href{https://github.com/lkeab/gaussian-grouping}{\faExternalLink}\\
         ReGS (\citeyear{mei2024regs})& 3DGS(\citeyear{kerbl20233d}), any 2D edit& 3DGS & image(s) &bounded/unbounded & single-style & appearance & \xmark \\ %Neurips24 
         % & & & & & & & \\
         \bottomrule
    \end{tabular}
    }
    \caption{Summary of selected neural field stylization methods. 3D Repr., Struct., geo., app. refer to 3D representation, data structure, geometry and appearance, respectively.  }
  
    \label{tab:sec_neural_field_style}\label{tab:sec_neural_field_tech_comp}
\end{table*}

\subsection{Neural Field-based Stylization} \label{sec:3d_style_neural_field}
A {neural field} is ``\textit{a field that is parameterized fully or in part by a neural network}"~\citep{po2023survey}. The advanced 3D representations of neural fields, especially neural radiance fields (NeRFs) \citep{mildenhall2020nerf, sun2022direct, muller2022instant, kerbl20233d}, store scene geometry and appearance in a neural network or explicit data structure, enabling photorealistic rendering and 3D stylization in the latent space. 
Compared to mesh models (Fig.~\ref{fig:pipelines}), which rely on texture maps to store various visual information like albedo, roughness, metalness, baked lighting, etc., neural fields store learned features that are mapped to an RGB image during rendering. 
Therefore, we can either stylize novel views during rendering without modifying the original neural field (Sec.~\ref{sec:nerf_nv_style}), or stylize the neural field itself by updating the stored latent features (Sec.~\ref{sec:nerf_stylize}). 
Methods that stylize novel views learn a universal style transformation module for 3D-aware view features, thus avoiding additional training for each input style instance. 
While the other approaches that stylize neural fields require optimization for every single style or input reference set, the stylized neural field assets allow regular neural rendering, thus supporting seamless usage in related tools and software.
Table \ref{tab:sec_neural_field_style} summarizes neural field-based stylization works, in terms of taxonomy and technical comparison. Please refer to the related surveys for a comprehensive review of neural fields and their applications \citep{xie2022neural,chen2024survey}.

\subsubsection{Feed-Forward Novel View Stylization} \label{sec:nerf_nv_style}
% LSNV
A straightforward approach to stylize a 3D scene is to stylize its novel views~\citep{huang2021learning}. However, it is known that a simple combination of existing 2D stylization and novel view synthesis methods would lead to blurry and inconsistent results. 
Instead, LSNV~\citep{huang2021learning} proposed a feed-forward point cloud feature transformation model that first reconstructs a 3D point cloud by back-projecting the points in feature maps extracted from multi-view images following depth guidance, and then feeds these features into the transformation network (similar to LST (\citeyear{li2019learning}) in Sec.~\ref{sec:2d-image-guide}) to obtain stylized point cloud features, which can be decoded to render novel views.

\begin{figure}[tbp]
    \centering
    \includegraphics[width=\linewidth]{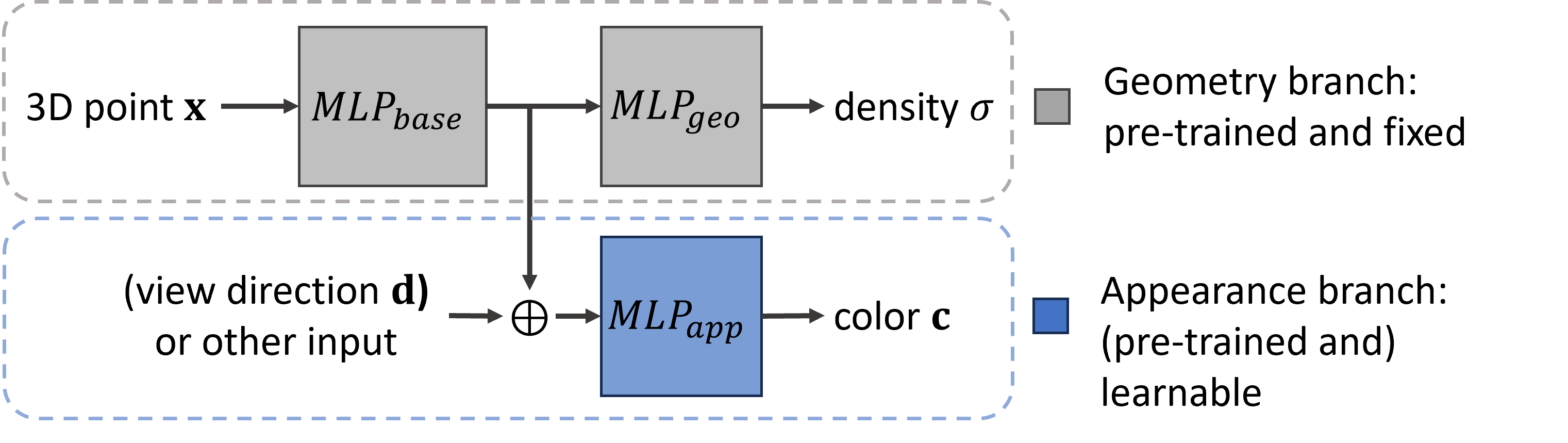}
    \caption{Canonical NeRF with geometry and appearance branches in 3D stylization.} 
    \label{fig:novel_view_nerf_branches}
\end{figure}

Similarly to neural style transfer, one can also perform arbitrary style transfer for novel views of a NeRF scene \citep{chiang2022stylizing, huang2022stylizednerf, liu2023stylerf, kim2024fprf}. This involves a two-phase process consisting of geometry reconstruction training (NeRF training) and {appearance stylization} training.
Chiang \etal (\citeyear{chiang2022stylizing}) proposed to transfer arbitrary artistic style to novel views on NeRF++ representation \citep{zhang2020nerf++} for large outdoor 360$\degree$ unbounded scenes. In the reconstruction phase, they separate geometry (density output) and appearance (view-dependent color output) into two branches, as illustrated in Fig.~\ref{fig:novel_view_nerf_branches}. In the stylization phase, they fix the geometry branch and use an MLP hypernetwork \citep{ha2016hypernetworks} with style features from a pre-trained VAE encoder to update the parameters of the appearance branch.
Since NeRF++ cannot support high-resolution image rendering at a fast speed, they propose to do small-patch sub-sampling \citep{schwarz2020graf} to compute content and style losses  \citep{huang2017arbitrary}.

As discussed in Sec.~\ref{sec:ArbiNST}, {AdaIN} and {LST} are two mainstream techniques for arbitrary style transfer, which are adapted in {StylizedNeRF}~\citep{huang2022stylizednerf} and {StyleRF}~\citep{liu2023stylerf} respectively.
{StylizedNeRF} applies a mutual learning strategy between a 2D arbitrary style transfer AdaIN \citep{huang2017arbitrary} model and a NeRF to achieve multi-view consistency and stylization of NeRF appearance. Specifically, they pre-train the style transfer model under the supervision of multi-view content, as well as style and additional consistency loss. During the mutual learning, they jointly train the style transfer model and a new appearance MLP branch with NeRF's fixed geometry branch. 

{StyleRF}~\citep{liu2023stylerf} employs a similar style transformation mechanism to {LSNV}, learning a 3D feature grid lifted from pre-trained VGG, and then applying linear style transformation on the weighted features of sampled points in the ray marching process of NeRF~\citep{chen2022tensorf}. Finally, a 2D CNN decoder is used to generate stylized views. {StyleRF} also conducts two-stage training, the stage of feature grid learning and reconstruction without viewing direction input (based on TensoRF \citep{chen2022tensorf}), and the stage of stylization training with fixed geometry.
Moreover, {StyleRF} illustrated its advantage of data-driven style training for style interpolation and multi-style transfer with a 3D mask.

Later works further explore arbitrary photorealistic style transfer~\citep{chen2022upst}, unbounded urban-scale scene style transfer~\citep{kim2024fprf} based on K-Planes representation and 2D-to-3D lifted DINO semantic features \citep{fridovich2023k, caron2021emerging}, and point cloud or mesh reconstruction of stylized novel views~\citep{ibrahimli2024muviecast}.

\subsubsection{Optimization-based Neural Field Stylization and Editing}~\label{sec:nerf_stylize}
In this section, we explore stylizing a neural radiance field (NeRF) by updating the scene information and features stored in neural networks or explicit data structure, rather than processing in the rendering phase.
Most NeRF-based approaches for appearance optimization follow a two-step procedure~\citep{zhang2022arf, fan2022unified, zhang2023ref, pang2023locally}. Initially, the scene's geometry and appearance are reconstructed from multiple posed views before stylization. Subsequently, during the optimization of the appearance style, the geometry can be either fixed or self-distilled during stylization, as depicted in Fig.~\ref{fig:novel_view_nerf_branches}. Notably, some methods incorporate geometry updates during the stylization phase~\citep{nguyen2022snerf, wang2023nerf, haque2023instructnerf}.

\begin{figure}[tbp]
    \centering
    \includegraphics[width=0.9\linewidth]{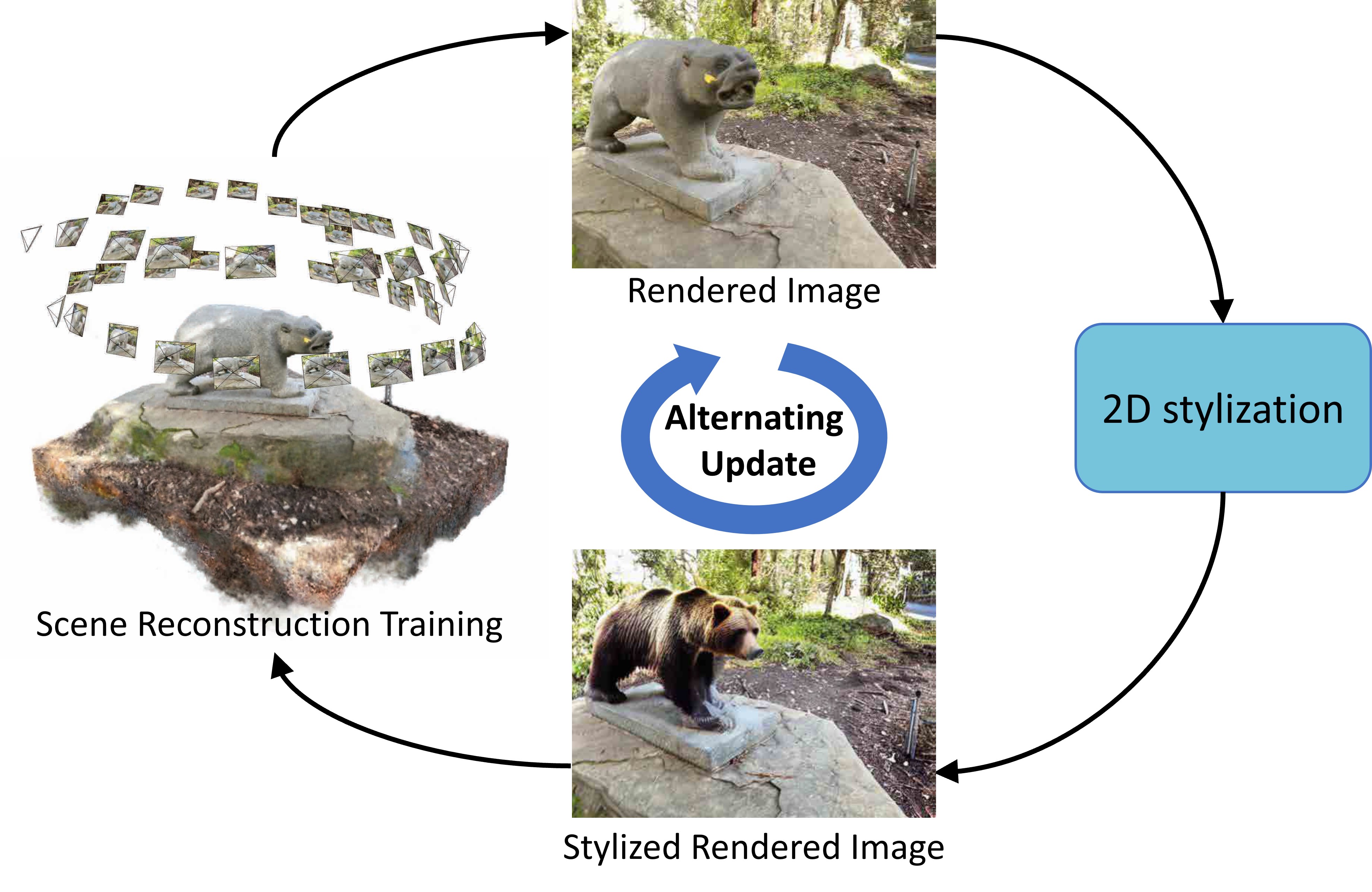}
    \caption{A general framework for NeRF stylization through alternating updates of multi-views~\citep{nguyen2022snerf, haque2023instructnerf}. Images adapted from \cite{haque2023instructnerf}.}
    \label{fig:novel_view_style_alterating}
\end{figure}

\paragraph{A General Optimization Framework} \label{sec:instructn2n}
\cite{nguyen2022snerf} proposed {SNeRF}, a general alternating optimization pipeline for novel view stylization by arbitrary off-the-shelf 2D style transfer methods and any NeRF methods (Fig.~\ref{fig:novel_view_style_alterating}). The proposed method follows a sequential process. Firstly, they reconstruct the NeRF scene with original content multi-views. Next, they iteratively stylize rendered multi-views and use stylized views to fine-tune the NeRF in a loop. Through several iterations, the entire NeRF representation gradually becomes 3D-aware stylized. 
% ICCV23 
\cite{haque2023instructnerf} further extended the framework for text-guided NeRF editing and introduced {Instruct-NeRF2NeRF}, which edits a NeRF scene by leveraging 2D diffusion priors. Similar to {SNeRF} (\citeyear{nguyen2022snerf}), they adopted an iterative process to update training views and the NeRF scene alternatively (Fig.~\ref{fig:novel_view_style_alterating}). They replace a training view by editing a rendered view from a training viewpoint using an off-the-shelf image-to-image diffusion model, e.g., Instruct-Pix2Pix \citep{brooks2023instructpix2pix}, then continue NeRF training on the updated training data.

Extensive experiments validated the flexibility of this general framework, showcasing its compatibility with off-the-shelf 2D style transfer methods across a range of NeRF or NeRF variants.
Despite promising results, this framework is limited by its time-consuming iterative nature and is vulnerable to variations in stylized views, which can lead to style dilution and inconsistency.
To address this issue, {ViCA-NeRF} \citep{dong2023vica} employs view-consistency-aware NeRF editing, which establishes explicit connections between different views and propagates the editing information from edited to unedited views.

\paragraph{Image-guided Radiance Field Stylization} \label{sec:arf}
Image-guided neural style transfer (Sec.~\ref{sec:2d-image-guide}) has undergone significant development over the years and has inspired a multitude of works focusing on image-guided NeRF stylization. Similarly, most works follow a two-step optimization process, including NeRF reconstruction and appearance stylization stages.

Notably, ARF~\citep{zhang2022arf} introduced the artistic radiance field approach, utilizing content loss \citep{gatys2016image} and nearest neighbor feature matching (NNFM) style loss (Eq.~\ref{eq:arf_nnfm_loss}) to optimize and stylize the appearance of a reconstructed NeRF scene with an exemplar style image. NNFM minimizes the cosine distance on VGG features between style reference and rendered image:
\begin{equation} \label{eq:arf_nnfm_loss}
    \mathcal{L}_{nnfm} = \frac{1}{N}\sum_{i,j}\min_{i',j'} D(F(cs)_{ij}, F(s)_{i'j'}),
\end{equation}
where $F(\cdot)_{ij}$ denotes the feature vector at pixel location $(i,j)$ of the feature map $F(\cdot)$.
Experiments validated NNFM loss in NeRF stylization yields more visually appealing results than typical Gram matrix loss \citep{gatys2016image} or CNNMRF loss \citep{li2016combining}.

However, {ARF} still lacks explicit semantic correspondences. To address this limitation, {Ref-NPR} \citep{zhang2023ref} proposed to first stylize a single view using structure-preserving 2D-stylization algorithms (Sec.~\ref{sec:2d-image-guide}) or manual editing, and then use this stylized reference view to construct a reference voxel dictionary for the scene, which enables matching of semantics and color features between edited view and the scene. The follow-up work CoARF~\citep{zhang2024coarf} allows for style transfer with precise control over specific objects indicated by 2D segmentation masks. These semantics are also used for calculating the NNFM loss. 
Concurrently, ReGS~\citep{mei2024regs} enables high-quality stylization that mimics reference textures while maintaining real-time rendering capabilities for free-view navigation. It achieves this by adapting 3D Gaussian Splatting (3DGS) and regularizing with scene depth.

Referring to the disentanglement of content and style representations for style transfer \citep{huang2018multimodal, lee2018diverse}, \cite{fan2022unified} proposed a generalizable model consisting of a style MLP and a content MLP to separately encode the style image and input scene, and an amalgamation MLP to output both final color and density, fusing style and content features. 
Additionally, \cite{pang2023locally} considered semantic style matching and added additional segmentation output in the geometry branch with an extra segmentation MLP after the hash encoding process of {iNGP}~\citep{muller2022instant}. 
Both works proposed to use conditional style representations by feeding a one-hot vector or style index to the neural field, enabling conditional stylization for several styles~\citep{fan2022unified, pang2023locally}. 

Apart from the artistic style transfer approaches, LipRF~\citep{zhang2023transforming} addressed the challenges in 3D photorealistic stylization by leveraging a Lipschitz MLP to transform the radiance appearance field during the stylization training stage. The scene views are first stylized by 2D photorealistic style transfer methods \citep{yoo2019photorealistic,wu2022ccpl} and then used to train the Lipschitz MLP. 

\paragraph{Text-guided Radiance Field Stylization} \label{sec:nerf_text_style}
The recent increasingly developed vision-language models and text-to-image diffusion models (Sec.~\ref{sec:2d-text-guide}) inspire the community to develop works on text-guided or text-to-image guided 3D scene stylization and editing \citep{wang2022clip, wang2023nerf, wang2023tsnerf, song2023blending, bao2023sine, haque2023instructnerf, sella2023vox, zhuang2023dreameditor, shum2023language}. Here we provide a brief discussion on some advances in text-guided NeRF stylization, showcasing their potential for rapid prototyping and customization of 3D asset designs.

% TVCG23
{NeRF-Art}~\citep{wang2023nerf} proposed text-guided NeRF stylization with profound semantics. Unlike simple color and shape stylization for objects in CLIP-NeRF \citep{wang2022clip} using a CLIP-based matching loss (Eq.~\ref{eq:clip_loss}), {NeRF-Art} realizes complex stylization on diverse shapes and scenes, such as turning a human face into a Tolkien elf. This method proposed to fine-tune the pre-trained NeRF \citep{yariv2021volume} using relatively direction CLIP loss (Eq.~\ref{eq:directional_clip_loss}), local and global contrastive CLIP-based loss~\citep{chen2020simple}, perceptual loss \citep{johnson2016perceptual} and a weight regularization \citep{barron2022mip} for sharper details. 
The follow-up work~\cite{wang2023tsnerf} further enhances the semantics-aware stylization with a semantic contrastive loss and fine-tuned CLIP with ArtBench artwork database \citep{liao2022artbench} for accurate artistic textual embedding.

Instead, {SINE} \citep{bao2023sine} employs a two-branch editing field to learn geometric and appearance adjustments. Given an edited view of a pre-trained NeRF scene, the method establishes the mesh prior using either DIF \citep{deng2021deformed} for specific object categories or using ARAP \citep{sorkine2007rigid} with depth estimation \citep{bhat2021adabins} and 2D feature matching \citep{jiang2021cotr}, and then composes the texture prior based on semantic features and structural self-similarity \citep{caron2021emerging, tumanyan2022splicing}. To preserve irrelevant areas, it distills a semantic feature field from DINO \citep{caron2021emerging} and clusters features in the edited region of the edited view.
To achieve precise manipulation of specific regions, {Blending-NeRF}~\citep{song2023blending} leverages CLIPSeg \citep{luddecke2022image}, a pre-trained image segmentation model, and employs a region loss for supervision.

To support editing, {DreamEditor}~\citep{zhuang2023dreameditor} directly uses 2D diffusion priors to enable precise editing while keeping irrelevant regions untouched. 
Specifically, this method transforms the NeRF into a mesh-based neural field by marching cubes \citep{lorensen1987marching} and distillation, with each mesh vertex assigned geometry and color features. 
To localize the edit regions aligned with a text prompt, the method leverages DreamBooth \citep{ruiz2023dreambooth} to fine-tune Stable Diffusion \citep{rombach2022high} using sampled views from a spherical viewing trajectory centered on the scene, and then retrieve 2D attention maps \citep{hertz2022prompt} as view masks that are later back-projected to 3D scene forming the 3D editing mask. Finally, geometry and color features, as well as mesh vertex positions in the 3D masked region are jointly optimized by the SDS loss (Sec. \ref{sec:3dgen}). Similar to implicit shape deformation methods \citep{bao2023sine, gao2023textdeformer}, {DreamEditor} employs mesh vertex regularizers, including Laplacian rigidity losses \citep{sumner2007embedded} among neighbor vertices, for smooth mesh deformation.

Facing the limitation of slow optimization in NeRF editing, {ED-NeRF}~\citep{park2024ed} proposed to edit a latent NeRF using 2D latent diffusion priors \citep{rombach2022high}, improving editing efficiency. However, multi-view rendering of latent features lacks geometry consistency. Therefore, they introduce a novel refinement layer with ResNet blocks and self-attention layers to refine inconsistent multi-view latent features. During editing, {ED-NeRF} employs delta denoising score (DDS) \citep{hertz2023delta} which is a difference between two SDS scores conditioned on two different text prompts, and a masked DDS for the target region segmented by CLIPSeg \citep{luddecke2022image} and SAM \citep{kirillov2023segment} to keep irrelevant regions unchanged.

Recent advancements in 3DGS-based scene editing offer new solutions for efficient optimization, fine-grained control, and high-quality scene segmentation. GaussCtrl~\citep{gaussctrl2024} and GaussianEditor~\citep{chen2023gaussianeditor} introduce text-driven editing for 3D Gaussian Splatting. GaussCtrl emphasizes multi-view consistency and depth-conditioned editing, while GaussianEditor enhances control and precision using Gaussian semantic tracing and hierarchical splatting. Gaussian Grouping~\citep{ye2025gaussian} extends Gaussian Splatting by incorporating identity encoding for object segmentation, enabling fine-grained scene understanding and versatile editing applications. These methods collectively enable real-time, high-quality, and efficient 3D scene manipulation across a wide range of applications, such as object removal, inpainting, and style transfer.

\begin{table*}[tbp]
    \centering
    \resizebox{\linewidth}{!}{ 
    \begin{tabular}{l ccccccc}
         \toprule
         Method &  Input & Guidance& Output & Scene & Optimization  & Style Genre & \faCode \\  
         
         \midrule
         % & & & & & & & \\
         TNST (\citeyear{kim2019transport}) & volume & image & volume & dynamic smoke & single-style & geometry, artistic & \href{https://github.com/byungsook/neural-flow-style/tree/tnst}{\faExternalLink}\\% (TNST) Transport-Based Neural Style Transfer for Smoke Simulations (SA 2019) 
         LNST (\citeyear{kim2020lagrangian}) & volume & image & volume & fluid  & single-style & geo.\&app., artistic & \href{https://github.com/byungsook/neural-flow-style/tree/lnst}{\faExternalLink}\\ % (LNST) Lagrangian Neural Style Transfer for Fluids (SG 2020) 
         SKPN (\citeyear{guo2021volumetric}) & volume & image & volume & dynamic/static model  & feed-forward & appearance & \xmark\\ % (SKPN) Volumetric Appearance Stylization with Stylizing Kernel Prediction Network (SG 2021)
         ENST (\citeyear{aurand2022efficient}) & volume & image & volume & smoke & multi-style, feed-forward & geometry & \xmark\\%(ENST) Efficient Neural Style Transfer for Volumetric Simulations (SA 2022) 
         \midrule
         PSNet (\citeyear{cao2020psnet}) & colored point cloud & image & colored point cloud & object & single-style & geo.\&app., photoreal & \href{https://github.com/xucao-42/psnet}{\faExternalLink}\\ %WACV 2020 
         PointInverter (\citeyear{kim2023pointinverter}) & point cloud & none & point cloud & object & feed-forward & geometry & \xmark\\ % 
         \midrule
         NeuralWavelet (\citeyear{hui2024wavelet}) & implicit shape & none & implicit shape & object & feed-forward & geometry & \href{https://github.com/edward1997104/Wavelet-Generation}{\faExternalLink}\\ %
         SPAGHETTI (\citeyear{hertz2022spaghetti}) & implicit shape & none & implicit shape & object & feed-forward & geometry & \href{https://github.com/amirhertz/spaghetti}{\faExternalLink}\\ %
         SALAD (\citeyear{koo2023salad}) & implicit shape & none, or text &  implicit shape & object & feed-forward & geometry & \href{https://github.com/KAIST-Geometric-AI-Group/SALAD}{\faExternalLink}\\ % 
         % & & & & & & & \\  
         \bottomrule
    \end{tabular}}
    \caption{Summary of selected neural stylization works for volume, point clouds and implicit shapes.}
    \label{tab:sec_volume_style}
\end{table*}

\subsection{Volume Stylization} \label{sec:3d_style_volume}
Compared to other representations, volume is an intuitive representation of 3D data, as naturally extended from 2D image representation. 
3D neural stylization can be performed on a volume representation, e.g., image-guided neural style transfer on volumetric simulation, particularly dynamic smoke \citep{kim2019transport} and fluids \citep{kim2020lagrangian}. Table \ref{tab:sec_volume_style} summarizes works for volumetric stylization.

In \cite{kim2019transport}, they use pre-trained Inception CNN model \citep{szegedy2016rethinking} as the single-view feature extractor and apply content loss and style loss (Sec.~\ref{sec:2d-image-guide}) for semantics-aware abstract style transfer. They proposed a transport-based neural style transfer ({TNST}) method on grid-based voxels to optimize a velocity field (i.e., voxel movement) from several multi-views from Poisson sampling around a small area of the trajectory, and use a differentiable smoke renderer to render grayscale images to represent pixel-wise volume density. For temporal consistency among frames during smoke simulation, they compose a linear combination of the recursive aligned velocity fields of neighbor frames for the velocity field of the current frame. Later in \citet{kim2020lagrangian}, they adopt particle-based attributes from mult-scale grids and optimize attributes of position, density, and color per particle, which intrinsically ensures better temporal consistency than recursive alignment of velocity fields \citep{kim2019transport}. It largely improves efficiency by directly smoothing density gradients in stylization from adjacent frames for temporal consistency, and by stylizing only keyframes and interpolating particle attributes in between.

Subsequently, one can use a feed-forward network to achieve fast volumetric stylization~\citep{aurand2022efficient}, reaching production-level quality~\citep{kanyuk2023singed, hoffman2023creating}. 
One can also learn an arbitrary appearance style transfer model for volumetric simulation via a volume autoencoder~\citep{guo2021volumetric}.

\subsection{Point Cloud Stylization} \label{sec:3d_style_point}

A {point cloud} is a discrete set of data points in 3D space, which may represent 3D shapes or objects. Each point can enclose additional attributes such as colors, normals \citep{pfister2000surfels} and spherical harmonic coefficients \citep{kerbl20233d} for rendering, or latent features \citep{huang2021learning} for 3D stylization. 

There are a few works that attempted stylization for point clouds (Table \ref{tab:sec_volume_style}). 
For example, {PSNet}~\citep{cao2020psnet} is a PointNet-based  \citep{qi2017pointnet} stylization network for point cloud color and geometry style transfer with a point cloud example or an image example. Similar to representing content and style features from a pre-trained model \citep{gatys2016image}, {PSNet} uses a PointNet-based classifier with two separate shared MLPs to extract intermediate outputs as geometry/content representation and regard the Gram-matrix of these outputs as appearance/style representation.
{PSNet} utilizes point-cloud-based content and style losses \citep{gatys2016image} to optimize the geometry and/or appearance color of the source point cloud, simply replacing VGG features with PSNet features. 
Since the Gram-based style representation is invariant to the number or the order of the input points, an example style image treated as a set of points can stylize source colored point cloud with only the target color style without shape deformation.
For point clouds generated by a generative model, one can learn to map a point cloud to the latent space for editing. 
{PointInverter} \citep{kim2023pointinverter} employed 3D point cloud GAN inversion and introduced an efficient way to conduct a 3D point cloud mapping to the latent space of a 3D GAN based on SP-GAN, a sphere-guided 3D point cloud generator \citep{li2021sp}. {PointInverter} resolves the point ordering issue during 3D point cloud inversion, while preserving point correspondences, which enables point editing in latent space.

\subsection{Implicit Shape Editing} \label{sec:3d_style_implicit_shape}
An {implicit primitive shape} is a 3D surface represented by an implicit distance function, such as a signed distance function (SDF) or truncated signed distance function (TSDF).
Powered by learning-based techniques, neural implicit shapes can represent 3D geometry as occupancy networks \citep{mescheder2019occupancy, peng2020convolutional}, distance fields \citep{park2019deepsdf}, volumetric radiance fields \citep{mildenhall2020nerf}, and Gaussian mixture models (GMMs).  
These neural continuous implicit representations have gained significant attention in the field of 3D shape generation and editing (Table \ref{tab:sec_volume_style}). 

Particularly, {NeuralWavelet} \citep{hui2024wavelet} utilized a compact wavelet representation consisting of coarse and detail coefficient volumes and designed a pair of diffusion generative models for coarse and detail 3D shape generation. During shape learning, an encoder is jointly trained to map the coarse coefficient volume to a condensed latent code. This latent code serves as a controllable condition for shape generation, inversion, and manipulation.
Recent approaches such as {SPAGHETTI} \citep{hertz2022spaghetti} and {SALAD} \citep{koo2023salad} adopted 3D generative models equipped with a hybrid representation that employs part-level disentanglement, extrinsic approximate shape and intrinsic geometric details disentanglement. In the hybrid representation, each part of the 3D shape is characterized by a set of extrinsic parameters, which form a Gaussian ellipsoid in 3D space (formulated by a 3D position with a covariance), capturing the approximate shape structure of that particular part. This part-level extrinsic-intrinsic disentanglement enables 3D shape generation and implicit shape manipulation such as local adjustment and part mixing. {SALAD} further incorporates text-guided shape part segmentation \citep{koo2022partglot} and performs text-guided shape completion.
These methods utilizing hybrid representations and incorporating text guidance offer promising advancements in 3D shape generation, editing, and manipulation, allowing for more intuitive and controlled transformations of 3D shapes.

\subsection{Practical Guidelines}
\label{sec:practical}
This section discusses practical aspects of 3D neural stylization methods, summarizing several design choices including 3D consistency, controllability, generalization, and efficiency.\\

\noindent\textbf{3D Consistency}. A particular challenge when performing stylization of 3D data is to ensure that view consistency is achieved so that the styles appear similar across views. We discuss common strategies to achieve view consistency, as follows. 

\noindent$\bullet$ \textit{View Sampling.} 
A reasonable camera sampling strategy is necessary for multi-view optimization without posed views.
A common strategy is to (randomly) sample around pre-defined principal cameras or along the camera trajectory, as summarized in Table~\ref{tab:3d_disc_view_plan}. \cite{michel2022text2mesh} devised a new training view selection scheme that samples view around an anchor view with the highest CLIP similarity with the target prompt.
Data augmentation is a common trick as well, such as random perspective transformation and random resize plus crop \citep{michel2022text2mesh, ma2023x, chen2024stylecity}, rendering with random backgrounds \citep{hwang2023text2scene}, mirroring and rotating subject elements \citep{aurand2022efficient}. 

\noindent$\bullet$ \textit{Constant Geometry.} Appearance-only 3D stylization requires keeping 3D geometry constant before and after stylization. The frequently used strategy is to fix geometry during the stylization stage.
Particularly, some neural field stylization works use hyper MLPs to predict stylized appearance branch parameters for stylization while fixing geometry branch parameters for 3D geometric consistency \citep{chiang2022stylizing, chen2022upst, wang2023tsnerf}.

\noindent$\bullet$ \textit{View-independent Appearance.} Some works of appearance stylization tend to maintain multi-view color consistency. However, some 3D representations may lead to multi-view appearance inconsistency, for example, view-dependent effects in radiance fields. To preserve multi-view color consistency, scenes are often optimized without viewing direction input in radiance fields, sacrificing view-dependent effects for better multi-view appearance consistency \citep{zhang2022arf, zhang2023ref, liu2023stylerf, pang2023locally}. 

\begin{table}[t]
    \renewcommand*{\arraystretch}{1.1}
    \setlength{\tabcolsep}{2pt}
    \centering
    \footnotesize
    \resizebox{\linewidth}{!}{ 
    \begin{tabular} {H l H c c c c}
         \toprule
         &Literature & Scene & Cam & \#View & Aug &  Sampling Criteria \\
         \midrule
         
         \multirow{8}{*}{\rotatebox{90}{\underline{Mesh}}} & \makecell[l]{Paparazzi \\(\citeyear{liu2018paparazzi})} & object & orth. & 1 & \xmark &
         \makecell[l]{uniform on offset surface,\\ face vertex normal} \\
         \cmidrule{2-7}
         & \makecell[l]{Text2Mesh\\ (\citeyear{michel2022text2mesh})} & object & pers. & 5$^a$ & \cmark &
         \makecell[l]{text-guided anchor view, \\sample around anchor}\\
         \cmidrule{2-7}
         & \makecell[l]{Text2Scene \\ (\citeyear{hwang2023text2scene})} & object & pers. & 20$^b$ & \cmark & \makecell[l]{random sample, \\evenly cover whole scene} \\ 
         \cmidrule{2-7} 
         & \makecell[l]{TEXTure \\(\citeyear{richardson2023texture})} & object & pers. & 8$^b$ & \xmark & \makecell[l]{around object \\ (incld. top/bottom)} \\
         \cmidrule{2-7}
         & \makecell[l]{Text2Tex \\(\citeyear{chen2023text2tex})} & object & pers. & 6+20$^{b,c}$ & \xmark & \makecell[l]{axis-align,\\face surface normal}\\
         \cmidrule{2-7}
         & \makecell[l]{StyleCity \\(\citeyear{chen2024stylecity})} & city & pers. & 1k+ & \cmark & \makecell[l]{look at several centroids, \\ sample around scene}\\
         \midrule
         
         \multirow{4}{*}{\rotatebox{90}{\underline{Volume}}} & \makecell[l]{TNST\\ (\citeyear{kim2019transport})}& object & orth. & 9 & \xmark & 
         \makecell[l]{look at object, \\sample around a path} \\
         \cmidrule{2-7} 
         & \makecell[l]{SKPN\\ (\citeyear{guo2021volumetric})} & object & pers. & 4 & \xmark & \makecell[l]{random sample in a path} \\
         \cmidrule{2-7} 
         & \makecell[l]{ENST\\ (\citeyear{aurand2022efficient})} & object & pers. & 1 & \cmark &  \makecell[l]{sample in a path} \\ 
         \bottomrule
    \end{tabular}
    }
    \begin{minipage}{\linewidth}
    \footnotesize{$^a$ total views equals views per iteration; \\$^b$ total views; \\$^c$6 views for generation, 20 out of 36 predefined views for refinement. }        
    \end{minipage}
    
    \caption{Summary of training view sampling in selected 3D neural stylization methods on object data. ``Cam" refers to camera type such as orthogonal or perspective camera. ``\#View" is the sample number of views for each iteration/frame, unless indicated otherwise. ``Aug" indicates rendered view augmentation. }
    \label{tab:3d_disc_view_plan}
\end{table}

\noindent$\bullet$ \textit{2D Priors.} With the growing popularity of large-scale pre-trained vision models including VGG, CLIP, DINO, and diffusion models \citep{simonyan2014very,radford2021learning,caron2021emerging,rombach2022high, zhang2023adding}, 3D-aware stylization can be achieved through multi-view optimization utilizing these 2D visual priors \citep{zhang2022arf, wang2023nerf, bao2023sine, haque2023instructnerf, Koo2024PDS}.

\noindent$\bullet$  \textit{3D Priors.} With numerous well-collected 3D datasets, pre-trained point cloud priors \citep{qi2017pointnet,li2021sp,nichol2022point} are popular to represent coarse geometry and facilitate geometric deformation among mesh, point cloud stylization works \citep{yin20213dstylenet, cao2020psnet, kim2023pointinverter}. Some other works leverage additional geometry guidance, such as depth map, normal map, and camera pose, for precise control of 3D-aware synthesis \citep{hollein2022stylemesh, yu2023texture, guo2023decorate3d}.

\noindent$\bullet$  \textit{Multi-view Attention.} Temporal attention mechanism is widely used in video generation methods~\citep{blattmann2023stable}, where an attention layer is applied to latent video frames to improve frame consistency. This concept can be adapted to the 3D domain. For example, VcEdit~\citep{wang2025view} inverse-renders the cross-attention maps from all views onto each Gaussian in the source 3DGS, creating an averaged 3D map. This 3D map is then rendered back to 2D, serving as the consolidated cross-attention maps for the originals, resulting in more coherent predictions.\\

\noindent\textbf{Controllability.} Stylization requires diversity and flexibility for users to design assets. We summarize some common strategies for different levels of control in 3D stylization. 

\noindent$\bullet$ \textit{Pre-trained Diffusion Models.} State-of-the-art diffusion models provide powerful controllability for content creation. For example, {TextureDreamer}~\citep{yeh2024texturedreamer} uses DreamBooth~\citep{ruiz2023dreambooth} to distill texture information from input reference images, and ControlNet~(Sec. \ref{sec:text_dif}, \citeyear{zhang2023adding}) to process additional 2D conditions, such as depth, normal, and edge maps. 

\noindent$\bullet$ \textit{Semantic Alignment.} Pre-trained segmentation models (e.g. Segment Anything~(\citeyear{kirillov2023segment})), semantic labels or descriptions can be integrated to empower semantics-aware stylization. Table \ref{tab:mesh_summary_semantic} shows some semantic matching tricks commonly used in 3D neural stylization. 
Some text-guided approaches rely on deliberate textual descriptions for local stylization \citep{michel2022text2mesh, ma2023x, wang2023tsnerf}, while some approaches consider explicit visual semantic matching such as {Text2Scene} with 3D label inputs \citep{hwang2023text2scene}.
\cite{gao2023textdeformer} proposed a regularization term to preserve identity for smooth deformation.
Some approaches consider using or lifting explicit semantic matching with off-the-shelf tools \citep{huang2022stylizednerf, zhang2023ref, pang2023locally, song2023blending, kim2024fprf}.
In addition, the reviewed works \citep{wang2023nerf, wang2023tsnerf} demonstrated that contrastive learning can effectively improve the learning of directional cues, such as textual semantics, in text-guided stylization.

\noindent$\bullet$ \textit{Explicit Representation.} An explicit scene representation allows much easier control and more precise manipulation. For example, \textit{DreamEditor}~\citep{zhuang2023dreameditor} transforms the NeRF into a mesh-based neural field with each mesh vertex assigned geometry and color features; \cite{chen2023gaussianeditor, hertz2022spaghetti, koo2023salad} use explicit 3D Gaussians for editing.

\noindent$\bullet$ \textit{3D Shape Inversion.} 
A commonly adopted technique for shape manipulation is 3D shape inversion, which inverts 3D representation into latent space learned by large-scale pre-trained 3D generative models \citep{nichol2022point, qi2017pointnet}. This approach enables shape editing in latent space and has been explored in various stylization works such as NeuralWavelet (\citeyear{hui2022wavelet}), SPAGHETTI (\citeyear{hertz2022spaghetti}), PointInverter (\citeyear{kim2023pointinverter}), and SALAD (\citeyear{koo2023salad}).\\

\begin{table}[t]
    \renewcommand*{\arraystretch}{1.1}
    \setlength{\tabcolsep}{2pt}
    \centering
    \footnotesize
    
    \resizebox{\linewidth}{!}{ 
    \begin{tabular}{m{.3\linewidth} H H m{.7\linewidth}}
         \toprule
          Literature & Scene & \centering Extra Input & \centering Semantic Matching Technique \tabularnewline
         \midrule
         \makecell[l]{Text2Mesh (\citeyear{michel2022text2mesh});\\ X-Mesh (\citeyear{ma2023x})} & \makecell[l]{object;\\ unbounded scene} &  \centering text & Design a text prompt for part-level localization and stylization. \\
         \hline
         TextDeformer (\citeyear{gao2023textdeformer}) & object &  \centering none & Use the identity-preserving term to keep deformation not far from input mesh.\\
         \hline
         Text2Scene (\citeyear{hwang2023text2scene}) & room &  \centering mesh label & 3D instance mesh label for instance-level stylization. \\
          \hline
         TSNeRF (\citeyear{wang2023tsnerf}) & \makecell[l]{object;\\ unbounded scene} & \centering text & Design a text prompt for part-level localization and stylization. \\
         \hline
         StyleRF (\citeyear{liu2023stylerf}) & \makecell[l]{bounded scene} & \centering none & Obtain 3D mask by NeRF-based object segmentation.
         \\
         \hline
         Ref-NPR (\citeyear{zhang2023ref}) & \makecell[l]{bounded or\\ unbounded scene} & \centering none & 
         Semantic matched edited view as reference; Propagate style features to other views.
         \\
         \hline
         LocalStyleNeRF (\citeyear{pang2023locally}) & \makecell[l]{room} & \centering seg maps & Train NeRF to obtain 3D segmentation from multi-view segmentation maps.
         \\
         \hline
         SINE (\citeyear{bao2023sine}) & & \centering none & Extract edited view's semantic features; Compute pixel-wise feature distance between edited region and training views. \\
         \hline         
         ED-NeRF(\citeyear{park2024ed}); BlendingNeRF (\citeyear{song2023blending})  & object & \centering text & Off-the-shelf text-guided segmentation model to segment each view for semantic-aware training. \\
         \hline
         DreamEditor (\citeyear{zhuang2023dreameditor}) & & \centering text & 2D attention maps in Diffusion as view masks, which are projected to 3D.\\
         \bottomrule
    \end{tabular}
    }
    
    \caption{Summary of semantic alignment in selected 3D neural stylization methods.}
    \label{tab:nerf_summary_semantic}
    \label{tab:mesh_summary_semantic}
\end{table}
\noindent\textbf{Efficiency.} Efficiency in 3D stylization is influenced by various factors such as style optimization methods and 3D representations. To enhance efficiency and minimize resource consumption, we present several tricks for efficiency improvement.

\noindent$\bullet$ \textit{Optimize in Coarse-to-Fine Manner.} 
Instead of direct stylization in final high resolution, some works operate 3D style optimization in the low resolution and then apply decoding or super-resolution techniques for higher optimization efficiency. 
For example, \cite{cao2023texfusion, knodt2023consistent} optimize latent features for 3D consistency and efficient diffusion. \cite{guo2023decorate3d, wu2023hyperdreamer} employ super-resolution models and \cite{yu2023texture,zeng2023paint3d} use texture refinement models to obtain high-quality outputs.

\noindent$\bullet$ \textit{Scene Representations.} For neural fields, there are various advanced representations for fast training or rendering, such as Plenoxels (\citeyear{yu_and_fridovichkeil2021plenoxels}), SNeRG~(\citeyear{hedman2021baking}), iNGP (\citeyear{muller2022instant}), DVGO (\citeyear{sun2022direct}), TensoRF (\citeyear{chen2022tensorf}), MobileNeRF (\citeyear{chen2023mobilenerf}) and 3DGS~(\citeyear{chen2024survey}). 
Table \ref{tab:sec_neural_field_tech_comp} features neural field stylization methods with applied base reconstruction techniques. Some 3D representations also tend to use advanced neural fields to represent style features such as neural style fields for meshes in \cite{michel2022text2mesh, ma2023x}.

\noindent$\bullet$ \textit{Optimize Rendering and Back-propagation.} 
In NeRF stylization, naive NeRF-based rendering is memory-intensive for a bulk of ray samplings and point queries, but style losses are based on full images. Therefore, some use patch-based training \citep{chiang2022stylizing}, deferred gradient back-propagation \citep{zhang2022arf}, and some separate forward and back-forward steps with full-res computation and patch-wise back-propagation \citep{zhang2023transforming, wang2023nerf, wang2023tsnerf}.

\noindent$\bullet$ \textit{Feed-forward Networks.} Instead of optimizing scene representation parameters, some works used feed-forward networks for efficient training/inference \citep{ma2023x, huang2021learning, chen2024pnesm, aurand2022efficient}.\\

\noindent\textbf{Generalization.} Stylization techniques across different scenarios and datasets are crucial for real-world applications, such as gaming or entertainment industries. We discuss some designs for generalization here.

\noindent$\bullet$ \textit{Universal Style Transfer Module.} 
Data-driven 3D neural stylization models train a universal 3D style transfer module that can generalize to new styles in a zero-shot manner. This type of work usually operates on novel view rendering rather than optimizing the scene features, as discussed in Sec.~\ref{sec:nerf_nv_style}.

\noindent$\bullet$ \textit{General Optimization Framework.} 
As presented in Sec.~\ref{sec:instructn2n}, {SNeRF} \citep{nguyen2022snerf} and {Instruct-NeRF2NeRF} \citep{haque2023instructnerf} introduced a general framework for a single scene optimization with either image or textual reference. They apply to any scene, any style, either geometry or appearance stylization.

% summarize datasets
\begin{table*}[!t]
    \centering
    \renewcommand*{\arraystretch}{1.2}
    \setlength{\tabcolsep}{2pt}
    \footnotesize
    \resizebox{\linewidth}{!}{ 
    \begin{tabular}{l l  |c c H c c c} 
    \toprule
    & Dataset Name &  Content & \#Scenes or Images  & \#Images & Synthetic/Real  & Highlights & \faCode \\
    \midrule
    \multirow{7}{*}{\rotatebox{90}{\footnotesize \underline{Mesh}}} 
     & TurboSquid (\citeyear{turbosquid_2023}) & (textured) mesh model & N/A & & synthetic & Professional publicly shared 3D models&\href{https://www.turbosquid.com/}{\faExternalLink} \\
     & CGTrader (\citeyear{cgtrader_2023})& (textured) mesh model & 1.81M & & synthetic & Professional publicly shared 3D models&\href{https://www.cgtrader.com/}{\faExternalLink}\\
     % & RenderPeople \citep{} & textured mesh model & 4.5K & & real & Scanned 3D people models \\
     & ShapeNet (\citeyear{shapenet2015}) & mesh model & 3M & & synthetic & Annotated 3D CAD object models&\href{https://shapenet.org/}{\faExternalLink} \\
     & COSEG (\citeyear{sidi2011unsupervised}) & mesh model & 172 & & synthetic & Shapes with segment annotations&\href{https://github.com/Ideas-Laboratory/shape-coseg-dataset}{\faExternalLink} \\
     & ModelNet (\citeyear{wu20153d}) & mesh model & 151,128& & synthetic &  3D CAD object models&\href{https://modelnet.cs.princeton.edu/}{\faExternalLink}\\
     & Thingi10K (\citeyear{zhou2016thingi10k}) & mesh model & 10K & & synthetic & 3D-printing models&\href{https://ten-thousand-models.appspot.com/}{\faExternalLink} \\
     & \makecell[l]{Objaverse 1.0/XL \\(\citeyear{deitke2023objaverse,objaverseXL})}& (textured) mesh model & 800K/10M$+$ & & synthetic & Annotated 3D objects and scenes sourced from Sketchfab&\href{https://objaverse.allenai.org/}{\faExternalLink}\\
     \midrule
     \rotatebox{90}{\footnotesize \underline{Pt.}}
     & DensePoint (\citeyear{cao2019point}) & colored point cloud & 10,454 & & synthetic & Point clouds built on ShapeNet and ShapeNetPart (\citeyear{yi2016scalable})&\href{https://github.com/qjadud1994/Point_Cloud_Colorization-pytorch}{\faExternalLink} \\
          \midrule
    \multirow{7}{*}{\rotatebox{90}{\footnotesize \underline{View Synthesis}}}
     & DTU (MVS) (\citeyear{jensen2014large}) & multi-views, MVS & 124 & & real & Scanned objects&\href{https://roboimagedata.compute.dtu.dk/?page_id=36}{\faExternalLink}\\
     & LLFF (\citeyear{mildenhall2019local}) & multi-views & 24 & & real & Forward-facing scenes&\href{https://drive.google.com/drive/folders/1QhpVUIjp9kwuCXTMqUC3jMYZFY4QPcHU}{\faExternalLink}\\
     & NeRF-Real (\citeyear{mildenhall2020nerf}) & multi-views & 2 & approx.100 & real &  Centered objects with background 360$\degree$ images&\href{https://drive.google.com/drive/folders/128yBriW1IG_3NJ5Rp7APSTZsJqdJdfc1}{\faExternalLink} \\
     & NeRF-Synthetic (\citeyear{mildenhall2020nerf}) & posed multi-views  & 8 & approx. 200 & synthetic & Centered objects w/o background 360$\degree$ images rendered by Blender&\href{https://drive.google.com/drive/folders/128yBriW1IG_3NJ5Rp7APSTZsJqdJdfc1}{\faExternalLink} \\
     & Tanks and Temples (\citeyear{knapitsch2017tanks})& multi-views & 21 & 147K & real & Large indoor and outdoor scenes&\href{https://www.tanksandtemples.org/}{\faExternalLink}\\
     & Real Lego (\citeyear{yu_and_fridovichkeil2021plenoxels}) & multi-views & 1 & 120 & real & Centered Lego with background 360$\degree$ images&\href{https://github.com/sxyu/svox2}{\faExternalLink}\\
     & Mip-NeRF 360 (\citeyear{barron2022mip})& multi-views & 9 & & real & Inward-facing object-centric 360$\degree$ indoor and outdoor scenes&\href{https://jonbarron.info/mipnerf360/}{\faExternalLink}\\
           \midrule
     \multirow{3}{*}{\rotatebox{90}{\footnotesize \underline{Style}}}
     & WikiArt (\citeyear{kaggle_2016})& painting & 79,433/23,815$^a$ &23,815 & real & Artistic paintings from WikiArt&\href{https://www.kaggle.com/c/painter-by-numbers}{\faExternalLink}\\ %  Painter by Numbers a.k.a Wikiart 
     % & New Wikiart\citep{artgan2018}&painting &N/A &10,628 & &\\
     & DPST (\citeyear{luan2017deep}) & photo & 60 & 60 & real & Photorealistic stylish photos&\href{https://github.com/luanfujun/deep-photo-styletransfer}{\faExternalLink}\\
     & COCO (\citeyear{lin2014microsoft}) & photo & 83K/41K$^a$ & 41,000 & real & Large-scale objects in context&\href{https://cocodataset.org}{\faExternalLink}\\
    \bottomrule
    \end{tabular}}
    \caption{Popular datasets for performance evaluation on 3D neural stylization. Pt. is the abbreviation of point cloud. $^a$Train/Test sets.}    \label{tab:datasets}
\end{table*}
\section{Datasets and Evaluation}\label{sec:eva}
In this section, we summarize the frequently used datasets for 3D neural stylization, introduce existing evaluation metrics and criteria for 2D and 3D stylization, and provide a benchmark of state-of-the-art 3D neural stylization works as the reference for future works. 

\subsection{Datasets} 
Datasets are essential for effective training and thorough validating of 3D stylization models in terms of applicable scenarios, stylization diversity, etc. Table~\ref{tab:datasets} illustrates selected popular 3D and 2D datasets for the evaluation of 3D neural stylization works, identifying their modality, scale, and other noteworthy features.

\subsection{Criteria and Metrics} \label{sec:eval_metrics}

The stylization and evaluation of 3D assets are commonly conducted through multi-view renderings, which could be attributed to the inherent way how people perceive and process 3D stuff, and the advancement of 2D large pre-trained vision models.
It is also possible to conduct stylization and evaluation directly in 3D space, mainly for the point cloud representation. For instance, 3DStyleNet~\citep{yin20213dstylenet} utilizes L1-Chamfer distance to guide the 3D shape transfer. PSNet~\citep{cao2020psnet} directly extracts style features from a point cloud using a modified PointNet~\citep{qi2017pointnet} structure. SpiceE~\citep{sella2023spic} introduces point cloud input as 3D shape prior to 3D generation. \cite{achlioptas2023shapetalk} provides a summary for evaluation metrics of 3D shape transfer.

We derive several critical aspects below from the state-of-the-art 3D neural stylization works for evaluating 3D stylization performance.
Overall, the main consideration includes the alignment with the target style, the preservation of the original content, the consistency between different views, the visual quality of the stylized results, and the efficiency of training/inference. 

\noindent$\bullet$ \textbf{Style Similarity.} 
Stylization tasks are driven by guidance information (\textit{i.e.} style reference), mainly images and texts. For measuring style similarity between the reference image and the rendering output, Gram matrix loss and AdaIN loss (\textit{i.e.} MSE of the channel-wise mean and variance) that introduced in Sec.~\ref{sec:2d-image-guide} are heavily used.  When the style reference is given by textual prompt, the most popular choice of recent works is CLIPScore~\citep{hessel2021clipscore}, which quantifies the correspondence between the rendered image and the textual prompt.

\noindent$\bullet$ \textbf{Content Preservation.} 
Content preservation is achieved to different extents in stylization works. In some works~\citep{chiang2022stylizing, chen2022upst, wang2023tsnerf} the stylization is conducted only for appearance while the 3D geometry is locked, which dramatically eliminates the morphing of geometric content. 
Some other works~\citep{zhang2022arf, zhang2023ref} aim for a balance of stylization and content texture preservation by training exclusively with view-independent texture colors, hence showcasing multi-view color consistency.

\noindent$\bullet$ \textbf{Multi-view Consistency.} 
Explicit 3D representations inherently provide multi-view {geometry} consistency. To measure multi-view {appearance} consistency, some 3D stylization works refer to video temporal short-range and long-range consistency evaluation \citep{lai2018learning}, using warping difference error and the warped LPIPS, via optical flow estimation or depth estimation \citep{chiang2022stylizing, liu2023stylerf, nguyen2022snerf, huang2021learning, hollein2022stylemesh}. 
CLIP can also be applied to evaluate multi-view semantic consistency by encoding adjacent views to CLIP space as proposed in \cite{haque2023instructnerf, ma2023x}.

\noindent$\bullet$ \textbf{Visual Quality.} 
For image synthesis, we expect synthesized images to look natural and contain as few artifacts as possible. The {inception score} ({IS})\citep{salimans2016improved} is designed to measure the image quality and diversity of generated images. 
Another popular metric is {Frechet Inception Distance} ({FID}) \citep{heusel2017gans}, which compares the distribution of generated images with the distribution of real images. It works well to decide if generated images are similar to objects in the target domain. For instance, {TSNeRF} \citep{wang2023tsnerf} used FID to evaluate the distance between stylized rendered views and the target art database.

\noindent$\bullet$ \textbf{Robustness and Efficiency.} 
For model robustness, {Ref-NPR} \citep{zhang2023ref} proposed to measure {PSNR} between rendered results around a
specific view angle. It is not essential for general 3D neural stylization evaluation, but it can be taken as a robustness reference. 
Regarding efficiency, important factors include the training time, inference speed, memory usage, data accessibility, model size and usability.

\noindent$\bullet$ \textbf{User Study.} 
The above metrics do not necessarily reflect and align with human bias, especially for subjective factors such as naturalness and attractiveness. Therefore, conducting a user study is usually a suitable option. 
A typical user study involves steps including recruiting participants, preparing study materials and questionnaires, collecting answers, and analyzing statistics. 
In 3D stylization, the most frequently evaluated metrics are ``stylization quality" and ``temporal consistency"~\citep{huang2021learning, chiang2022stylizing, chen2022upst, liu2023stylerf}.

\subsection{Benchmark of 3D Stylization} \label{sec:eval_benchmark}

In this section, we provide a benchmark evaluation in Table~\ref{tab:eva_whole} of state-of-the-art mesh-based and neural field-based neural stylization methods in terms of the criteria discussed above, followed by a high-level discussion of the insights gained. The methods can be categorized into text-guided or image-guided object mesh texture stylization, text-guided neural field stylization, and image-guided neural field artistic stylization.

\begin{table*}[!t]
    \centering
    \renewcommand*{\arraystretch}{1.2}
    \setlength{\tabcolsep}{1.5pt}
    \footnotesize
    \resizebox{\linewidth}{!}{
    \begin{tabular}{c l c c c c c c c c }
        \toprule
        3D Repr. & Method & Guidance & Optimization  & \makecell{Style Similarity \\ (ClipScore$\uparrow$)} & \makecell{Multi-view Consistency \\ (ClipVar$\uparrow$)} & \makecell{Visual Quality \\ (FID$\downarrow$)} & \makecell{GPU\\(GiB)} & \makecell{Pre-training Time\\(min)} & \makecell{Optimization Time\\(min)} \\
        \midrule
        \multirow{7}{*}{Mesh} 
         & TEXTure (\citeyear{richardson2023texture}) & \multirow{6}{*}{text} & single-style  & \colorfirsttext{\textbf{28.43}} & 81.39 & 201.77 & $\sim 11$ & No need & $\sim 1.4$ \\
         & Paint-it (\citeyear{youwang2023paint})& & single-style  & 27.87 & 81.77 & 164.08 & $\sim 30$ & No need & $\sim 30$ \\
         & Paint3D (\citeyear{zeng2023paint3d}) && single-style  & 27.54 & 80.41 & 167.13 & $\sim 25$ & No need & $\sim 3$ \\
         & TexPainter (\citeyear{zhang2024texpainter}) && single-style  & 26.73 & 82.62 & 157.92 & $\sim 14$ & No need & $\sim 6$ \\
         & FlashTex (w/ depth) && single-style  & 27.40 & \colorfirsttext{\textbf{84.36}} & 162.99 & $\sim 14$ & No need & $\sim 2.9$ \\
         & FlashTex (\citeyear{deng2024flashtex}) && single-style & 25.73 & 83.70 & \colorfirsttext{\textbf{155.73}} & $\sim 16$ & - & $\sim 4.8$ \\
        \bottomrule
        \\
        \toprule
        3D Repr. & Method & Guidance  & Optimization & \makecell{Style Similarity \\ (ClipScore$\uparrow$)\\in-domain/out-domain} & \makecell{Multi-view Consistency \\ (ClipVar$\uparrow$)\\in-domain/out-domain} & \makecell{Visual Quality \\ (FID$\downarrow$)} & \makecell{GPU\\(GiB)} & \makecell{Pre-training Time\\(min)} & \makecell{Optimization Time\\(min)} \\
        \midrule
        \multirow{3}{*}{Mesh} 
         & TEXTure (\citeyear{richardson2023texture}) & \multirow{3}{*}{\makecell{concrete\\ object image}} & single-style  & 79.70/64.66 & 86.07/\colorfirsttext{\textbf{88.49}} & 172.44 & $\sim 11.5$ & $\sim 140$ & $\sim 0.9$ \\
         & Paint3D (\citeyear{zeng2023paint3d}) && single-style   & \colorfirsttext{\textbf{81.54}}/70.23 & \colorfirsttext{\textbf{86.77}}/85.51 &  133.01 & $\sim 26$ & No need & $\sim 3.1$ \\
         & Easi-Tex (\citeyear{perla2024easitex}) && single-style   & 81.37/\colorfirsttext{\textbf{70.50}} & 84.75/83.50 & \colorfirsttext{\textbf{120.63}} & $\sim 12$ & No need & $\sim 15.4$ \\
        \bottomrule
        \\
    
        \toprule
        3D Repr. & Method & Guidance & Optimization & \makecell{Style Similarity \\ (ClipScore$\uparrow$)} & \makecell{Multi-view Consistency \\ (ClipVar$\uparrow$)} & \makecell{Visual Quality \\ (FID$\downarrow$)} & \makecell{GPU\\(GiB)} & \makecell{Pre-training Time\\(min)} & \makecell{Optimization Time\\(min)} \\
        \midrule
        \multirow{4}{*}{Neural Field}  
         & InstructN2N  (\citeyear{haque2023instructnerf}) & \multirow{4}{*}{text} & single-style & \colorfirsttext{\textbf{26.47}} & 87.43 & - & $\sim 13$ & $\sim 25$ & $\sim 60$ \\
         & VICA-NeRF (\citeyear{dong2023vica}) && single-style   & 25.22 & 84.97 & - & $\sim 7.7$ & $\sim 25$ & $\sim 82$ \\
         & InstructGS2GS  (\citeyear{igs2gs})&& single-style   & 22.16 & 86.26 & - & $\sim 4$ & $\sim 5$ & $\sim 18$ \\
         & PDS (\citeyear{Koo2024PDS}) && single-style   & 24.24 & \colorfirsttext{\textbf{87.63}} & - & $\sim 16$ & $\sim 5$ & $\sim 200$ \\
        \bottomrule
        \\
    
        \toprule
        3D Repr. & Method & Guidance & Optimization & \makecell{Style Similarity \\ (Gram Loss$\downarrow$$\times 1e5$)} & \makecell{Multi-view Consistency \\ (LPIPS$\downarrow$$\times 1e2$)\\short-range/long-range} & \makecell{Visual Quality \\ (FID$\downarrow$)} & \makecell{GPU\\(GiB)} & \makecell{Pre-training Time\\(min)} & \makecell{Optimization Time\\(min)} \\
        \midrule
        \multirow{8}{*}{Neural Field}  
         & ARF (\citeyear{zhang2022arf}) & \multirow{8}{*}{\makecell{artistic \\style image}} & single-style   & 6.25 & 7.52/\colorfirsttext{\textbf{4.60}} & - &  $\sim 14$ &  $\sim 20$ &  $\sim 20$\\
         & SNeRF-G (\citeyear{nguyen2022snerf}) && single-style   & \colorfirsttext{\textbf{3.12}} & 11.72/4.62 & - & $\sim 14$  & $\sim 20$ & $\sim 100$\\
         & INS (\citeyear{fan2022unified}) && single-style   & 4.78 & 11.15/8.12 & - & $\sim 24$ & $\sim 2.3\times1e3$ & $\sim 120$ \\
         & Ref-NPR (\citeyear{zhang2023ref}) && single-style   & 4.47 & 8.02/5.30 &  - & $\sim 14$  & $\sim 20$ & $\sim 20$ \\
          & LSNV (\citeyear{huang2021learning}) && feed-forward  & 9.09 & 8.84/8.05 & \colorfirsttext{\textbf{247.29}} & $\sim 12$ & $\sim 1.7\times1e3$ & - \\
         & HyperStyle (\citeyear{chiang2022stylizing}) && feed-forward   & 5.41 & 11.23/6.28 & 289.29 & $\sim 72$  &  $\sim 460$ & - \\
         & StyleRF (\citeyear{liu2023stylerf}) && feed-forward   & 4.84 & 6.12/5.42 & 267.11 &  $\sim 24$  & $\sim 300$ & -  \\
         & StyleGaussian (\citeyear{liu2023stylegaussian}) && feed-forward   & 5.08 &  \colorfirsttext{\textbf{5.51}}/8.02 &  293.46 & $\sim 19$ & $\sim 350$ & - \\
        \bottomrule
    \end{tabular}
    }
    \caption{Evaluation of selected works across 3D representation and guidance type. w/ depth stands for with pre-trained depth-ControlNet. SNeRF-G is reproduced by Plenoxels (\citeyear{yu_and_fridovichkeil2021plenoxels}) and \cite{gatys2016image}.}   \label{tab:eva_whole}
\end{table*}

\subsubsection{Experimental Settings}

\noindent$\bullet$~\textbf{Datasets.} For mesh-based stylization methods with image guidance, we create 300 object-image pairs from Objaverse~\citep{deitke2023objaverse} dataset: 100 objects with their own rendered images, 100 with rendered images of other objects in the same category, and 100 with rendered images from different categories. The first two parts aim to evaluate the capacity of ``in-domain'' texture transfer, while the last part tests the capacity of ``out-of-domain'' texture transfer. All the images are rendered in $2048\times2048$ resolution. For TEXTure~\citep{richardson2023texture}, we fine-tuned ten diffusion models following their official requirement to conduct the texture transfer.
For mesh-based stylization methods with text guidance, we use the same 100 selected objects from Objaverse and directly use the object name to construct the text prompt. 

For neural field-based stylization methods with image guidance, we include eight scenes from three public datasets, including single-object scenes (chair, mic) in NeRF-Synthetic \citep{mildenhall2020nerf} that are inward-facing 360$\degree$ objects without background, forward-facing real scenes (fern, flower, horns, trex) in LLFF dataset \citep{mildenhall2019local}, and large unbounded real scenes (Truck, Playground) in Tanks\&Temples dataset \citep{knapitsch2017tanks}. Particularly, masked large scenes Caterpillar and Truck without background \citep{knapitsch2017tanks} are used instead for {StyleRF} and {INS}. The style reference images are selected from WikiArt~\citep{kaggle_2016} dataset. 120 WikiArt~\citep{kaggle_2016} style references are used for feed-forward methods, and 6 WikiArt images for single-style optimization methods. 
We select artistic images here because neural field-based methods usually have larger scenes with multi-objects and backgrounds, and single-object images won't lead to satisfying results. Conversely, such artistic images with abstract concepts tend to ignore the detailed semantics of a concrete object, and thus are not suitable for most mesh stylization methods that focus on a single object.
For neural field-based stylization methods with text guidance, we select two unbounded and two forward-facing scenes (farm, campsite; fangzhou, person) from InstructN2N~\citep{haque2023instructnerf} and test four style prompts for each of them.

\noindent$\bullet$~\textbf{Metrics.} For style similarity, we compute Gram Loss for artistic image-guided works and ClipScore for others.
FID~\citep{heusel2017gans} is adopted for measuring the visual quality of selected methods, using the rendered views of stylized results as evaluation samples and style reference images as ground truths. The original rendering views of selected objects from the Objaverse dataset are used as ground truths for evaluating text-guided mesh stylization works. Since the FID metric needs to be calculated on a large number of evaluation images, we skipped a few works that are hard to obtain a large amount of ground truth data~\citep{haque2023instructnerf, dong2023vica, igs2gs, Koo2024PDS} or have a relatively slow optimization speed which prevents generating a sufficient number of evaluation samples~\citep{fan2022unified, nguyen2022snerf, zhang2022arf, zhang2023ref}.

For multi-view consistency, we utilize CLIP-Var~\citep{li2025learning} to take the minimum value of cosine similarity between CLIP features of uniformly sampled views as a metric, which derives from the idea that images of the same object from multiple views have the same semantics. 
For the artistic style transfer task, we compute short-range and long-range warp error with masked LPIPS scores via off-the-shelf optical flow estimator RAFT \citep{teed2020raft}.

The GPU consumption, pre-training time and optimization time are measured on RTX 5880 Ada GPUs with 48GB memory per GPU. The pre-training time denotes the normal duration for required additional training of the method before conducting stylization (while the original authors may have provided trained weights), like training a ControlNet~\citep{deng2024flashtex}, training a feature transformation module~\citep{liu2023stylerf, chiang2022stylizing}, training the original 3D reconstruction, etc.
Please refer to our evaluation code repository for details: \url{https://github.com/chenyingshu/advances_3d_neural_stylization}.

\subsubsection{Discussion}
Through the theoretical analysis, benchmarking and practical experience, we aim to address a research question: \emph{How do various factors such as 3D representation, optimization methods, guidance, etc., impact stylization outcomes across different dimensions like visual quality, consistency, and efficiency?}  We will delve into this inquiry through the following key points.

\noindent$\bullet$~\textbf{Optimization - How to conduct efficient optimization?} When aiming for efficiency, effective strategies include employing large pre-trained models and training task-specific adapter modules. For example, TEXTure and FlashTex \citep{richardson2023texture, deng2024flashtex} can synthesize stylish texture in high quality in under five minutes by leveraging large pre-trained diffusion models as priors. Additionally, some utilize feed-forward processing to enhance efficiency during stylization inference, removing the necessity for per-style optimization, as demonstrated in works like StyleRF and StyleGaussian\citep{liu2023stylerf, liu2023stylegaussian}.

\noindent$\bullet$~\textbf{Guidance - How to provide effective guidance?}
In stylization, a visual prompt can efficiently convey intricate details, especially for complex designs or expectations that are hard to articulate in natural language. Conversely, textual prompts offer greater flexibility allowing for easy adjustments. 
In Table~\ref{tab:eva_whole}, image-guided mesh stylization methods \citep{richardson2023texture, zeng2023paint3d, perla2024easitex} exhibit higher CLIP-scores compared to text-guided approaches \citep{richardson2023texture, youwang2023paint, zhang2024texpainter, deng2024flashtex}.
This disparity stems from the calculation principle of CLIP-score that captures multi-concept features from the inputs and measures the similarity of the features, where image-guided texture transfer can directly reconstruct the features from the reference image and thus easily achieve higher CLIP-scores. Meticulous prompt engineering is required to achieve similar results using natural language. 

Beyond 2D guidance, 3D guidance proves effective for tasks like 3D shape transfer, often through the point cloud representation which enables 3D shape similarity calculation using metrics like Chamfer distance. The point cloud representation offers efficiency for physics simulation, scalability, and other advantages. 3D Gaussian Splatting~\citep{kerbl20233d} akin to point clouds has great potential in such topics \citep{kotovenko2024wast}.

\noindent$\bullet$~\textbf{Visual Quality - How to enrich visual effects while reducing artifacts?}
State-of-the-art 3D stylization methods improve visual quality by developing view-dependent appearances based on CG empirical models \citep{deng2024flashtex}, utilizing multiple vision priors \citep{haque2023instructnerf, youwang2023paint, zhang2022arf}, and data-driven learning \citep{huang2021learning, liu2023stylerf}.
For example, Easi-Tex~\citep{perla2024easitex} employs a pre-trained IP-Adapter~\citep{ye2023ip} and an edge ControlNet to faithfully extract the texture and shape details respectively, demonstrating both high visual quality and style similarity, even if there are significant discrepancies between the input and the reference object (we denote it as ``out-domain'' type in Table~\ref{tab:eva_whole}).
FlashTex~\citep{deng2024flashtex} trains a novel LightControlNet, which learns from numerous rendered images of objects with different materials and enables providing rich visual details in texture generation.

\noindent$\bullet$~\textbf{Consistency - How to ensure multi-view consistency?} 
Existing works strive to achieve multi-view consistency when rendering photorealistic or artistic 3D scenes. As mentioned in the practical guidelines (Sec.~\ref{sec:practical}), some works directly construct view-independent objects/scenes~\citep{zhang2022arf, liu2023stylerf, zeng2023paint3d} which largely alleviates the worry. However, it will significantly improve the overall quality to provide view-dependent effects or make the object/scene light-aware~\citep{zhang2024dreammat, deng2024flashtex}. Similar to the ideas of linking 2D to 3D stylization in Sec.~\ref{sec:link2d3d}, one way is to devise a dedicated loss item to enforce multi-view consistency~\citep{zhang2023ref, mei2024regs}. ARF~\citep{zhang2022arf} presents another way that applies a simple linear transformation of colors in RGB space for all rendered views to match the color statistics of the style image, which greatly improves the consistency between the rendered views. Last but not least, we can incorporate additional guidance (depth map, normal map, etc.) for generative models. As seen in Table~\ref{tab:eva_whole}, FlashTex~\citep{deng2024flashtex} with depth ControlNet achieves the highest multi-view consistency among the text-guided mesh stylization methods. Compared to Paint3D and Easi-Tex which only use one image as style reference, TEXTure achieves an overall higher Clip variance, probably benefiting from fine-tuning a diffusion model with multi-view renderings of the target object for texture transfer.

\noindent$\bullet$~\textbf{Scalability - How to adapt stylization to different sizes of scenes?} 
Mesh-based stylization is thriving in 3D object assets, which is suitable for benchmarking with accessible 3D object datasets. Some researchers also attempt to stylize room-scale \citep{chen2023scenetex, hollein2022stylemesh} or city-scale scenarios \citep{chen2024stylecity} concerning complex semantics and view sampling strategies. 
By employing neural fields, 3D representations support various scales of scenes, from naive NeRF and grid-based radiance fields for object-centric and feed-forward scenes \citep{fan2022unified, zhang2022arf, liu2023stylerf} to 3DGS for scenes in the wild \citep{igs2gs, liu2023stylegaussian}. Stylization techniques can be tailored to specific representations considering the structure types such as grids \citep{liu2023stylerf}, points \citep{huang2021learning}, and optimization strategies, as summarized in Table \ref{tab:sec_neural_field_style}.

\section{Applications}\label{sec:app}
\begin{figure*}
    \centering
    \includegraphics[width=\linewidth]{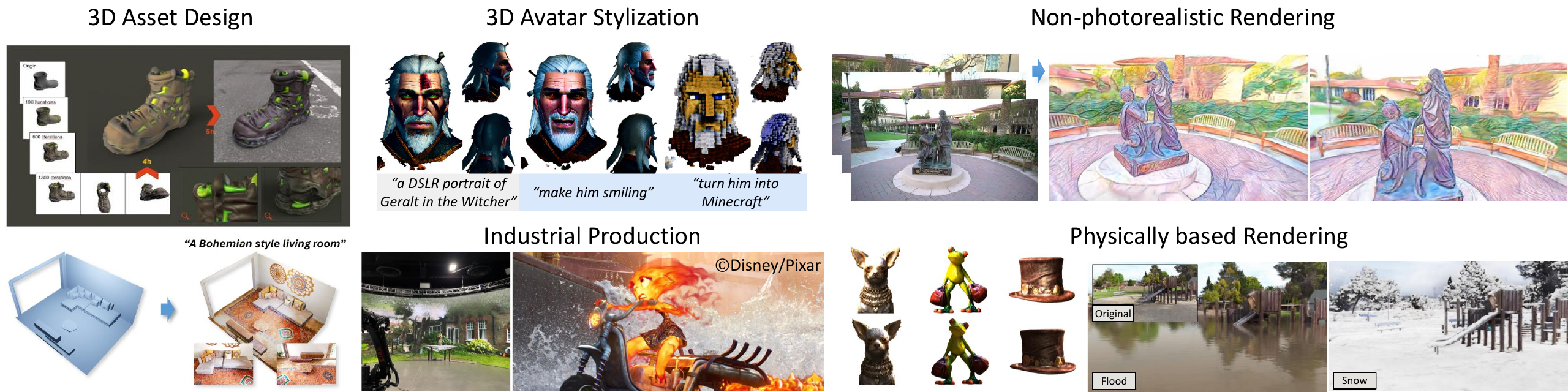}
    \caption{Applications of 3D neural stylization. Images adapted from \cite{endava20223dasset, chen2023scenetex, han2024headsculpt, volinga_2023, kanyuk2023singed, zhang2022arf, wu2023hyperdreamer, li2023climatenerf}.}
    \label{fig:applications}
\end{figure*}

% domain introduction + advantages + example works
The burgeoning technologies for generating and manipulating 3D assets are unleashing the power of creativity and revolutionizing the way that we perceive and interact with visual content. 
3D neural stylization sheds light on a new paradigm of providing infinite aesthetic possibilities from classic paintings to futuristic concepts, enhanced immersive experiences for virtual and augmented reality environments, seamless integration for cross-industry applications including advertising and marketing, fashion and product design, film and game development, architecture and environment visualization, and interactive education and learning, etc. Some examples are visualized in Fig.~\ref{fig:applications}.     
In this section, we present some representative and promising applications of 3D neural stylization.

\begin{figure}[t]
    \centering
    \includegraphics[width=\linewidth]{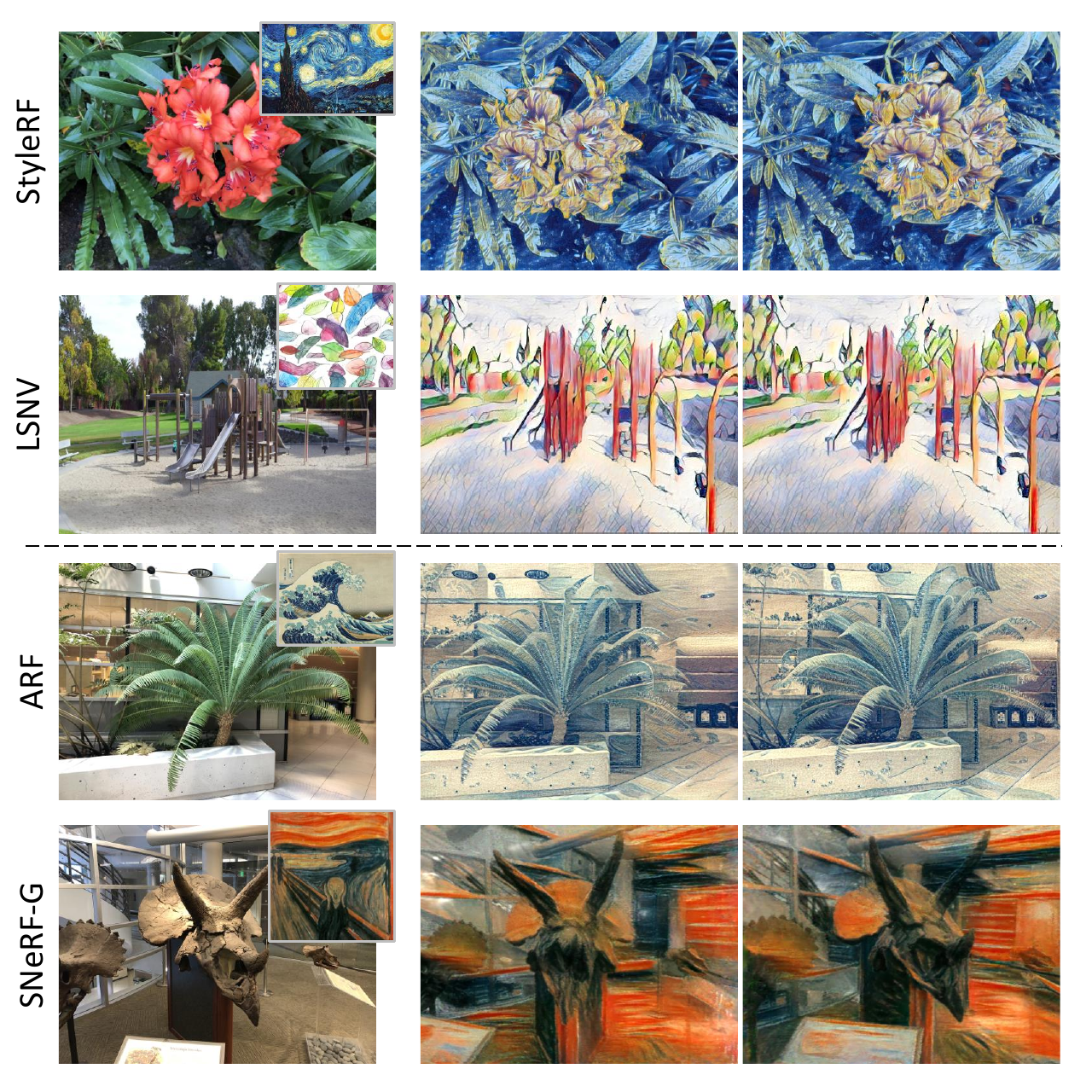}
    \caption{Neural field stylization results. Zoom in for details.}
    \label{fig:eval_results}
\end{figure}

\subsection{3D Asset Design}  % asset generation/improvement/ adapt rendering pipeline

3D asset design and modeling involve constructing shapes, textures, materials, etc.  Harnessing advanced neural stylization techniques, automatic 3D design becomes more flexible in a controllable way with text prompts, images or 3D references.

\noindent\textbf{Single Object Design.}
In the creation of consumer products, the stylization of a single object can enhance its market appeal. Neural stylization enables fast illustration of ideas for a more effective discussion among designers, developers and customers, especially in the prototyping phase.
Text-guided mesh stylization provides a flexible way to design 3D assets. For example, after the launch of {Text2Mesh} \citep{michel2022text2mesh}, digital R$\&$D designers and artists employed this tool for 3D gaming asset designs \citep{endava20223dasset} and artwork creation \citep{varvara2022psy}. 
Advanced techniques in automatic 3D shape mixing and morphing \citep{hui2022wavelet, gao2023textdeformer} are also promising for speedy 3D asset design and production.

\noindent\textbf{Room Decoration.}
Current digital room/house decoration tools support adding pre-made assets to a virtual scene and adjusting their places, or using the device's camera to map the room and place provided furniture~\citep{yu2011make,IKEAPlace,Houzz,Planner5D}. However, the asset library provides limited types and styles of 3D models, and the overall style of the whole space is neglected.
Recent works that leverage 3D neural stylization techniques explore more possibilities in room decoration.
{DreamSpace}~\citep{yang2023dreamspace} allows users to personalize the appearance of real-world scene reconstructions with text prompts, and delivers immersive VR experiences on HMD devices. SceneTex~\citep{chen2023scenetex} generates high-quality textures for 3D indoor scenes from the given text prompts, which provide a consistent stylization for the whole space. Instead of entire room stylization, {Text2Scene}~\citep{hwang2023text2scene} focuses on the stylization of individual object meshes in an indoor scene.

\subsection{3D Avatar Stylization}
Avatar stylization is a long-standing and popular research area, that enables interesting applications such as cartoonization for 2D or 3D-aware portraits \citep{jang2021stylecarigan, song2021agilegan, yang2022pastiche, zhang2023deformtoon3d}. With neural stylization techniques and novel 3D representations such as NeRF, there appear stylization solutions for 3D avatars \citep{perez2024styleavatar, zhang2024text, han2024headsculpt}.

For example, the general NeRF stylization framework {SNeRF} \citep{nguyen2022snerf} supports style transfer for dynamic NeRF avatars.
{3DFaceHybrid} \citep{feng20243d} achieves arbitrary style transfer for a NeRF-based face by lifting 2D pre-trained face style transfer knowledge \citep{yang2022pastiche} to the 3D face mesh. {StyleAvatar} \citep{perez2024styleavatar} enables either image- or text-guided stylization for animatable avatars from a phone scan \citep{cao2022authentic} with CLIP supervision. {TECA} \citep{zhang2024text} generates a detailed 3D avatar composition based on a given text description. The avatar includes a mesh-based face and body, and NeRF-based hair, clothing, and other accessories. {HeadSculpt} \citep{han2024headsculpt} generates and edits a 3D-consistent head avatar with text prompts via diffusion priors \citep{brooks2023instructpix2pix}, achieving editions such as realistic or artistic head generation, expression editing, cartoonization, etc.

\subsection{Non-photorealistic Rendering}
Compared to traditional non-photorealistic rendering techniques \citep{gooch1998non, gooch2001non} with low-level control of simple strokes and textures, neural stylization techniques realize general stylization for arbitrary style targets, offering high-level controllability with reference and semantics and higher speed for stylized assets production. 3D artistic stylization works have shown the potential of efficient NPR for a scene \citep{huang2021learning, chiang2022stylizing, huang2022stylizednerf, liu2023stylerf, zhang2022arf, fan2022unified, nguyen2022snerf, zhang2023ref, wang2023nerf, haque2023instructnerf}.

\subsection{Physically Based Rendering}
Physical properties in 3D scenes enable photorealistic rendering and editing.

\noindent\textbf{Texture Stylization.} {TANGO} \citep{lei2022tango} tends to optimize texture material parameters by CLIP supervision. It trains MLPs given the point and its normal to generate SVBRDF parameters and normal offset, which enables photorealistic rendering. Its follow-up work {3DStyle-Diffusion} \citep{yang20233dstyle} further incorporates depth-guided ControlNet \citep{zhang2023adding} for score distillation, enabling high-quality fine-grained texture stylization.

\noindent\textbf{3D Generation and Editing.} {HyperDreamer} \citep{wu2023hyperdreamer} achieves single-image-to-3D generation with physical decomposition including semantics, albedo, specular, roughness, and normal. It supports diverse downstream tasks such as relighting, text-guided and part-aware editing.
{Decorate3D} \citep{guo2023decorate3d} converts NeRF scene to mesh for geometry and material decomposition. The decoupled geometry and UV texture representations support controllable texture editing and generation with text instructions.

\noindent\textbf{Simulation.}
{ClimateNeRF} \citep{li2023climatenerf} fuses weather physical simulation with NeRF rendering to create NeRF scenes with realistic weather effects such as smog, snow, and floods. {PhysGaussian} \citep{xie2023physgaussian} integrates physics-based dynamics simulation, specifically the Material Point Method (MPM) simulation, to deform a 3DGS scene. By merging realistic rendering and physical simulation, these approaches have the potential to enhance the realism of virtual games and films.

\subsection{Industrial Production}
3D neural stylization provides automatic stylization techniques for 3D assets including mesh, point cloud, volumetric simulation, and novel views. Stylized assets can be seamlessly integrated into traditional computer graphics rendering pipelines and software, such as meshes with new stylized texture, re-colored point clouds, and stylized volumetric simulation. Implicit reconstructed scenes, such as NeRF, can be exported as textured mesh or rendered by game engine plugins such as Luma AI's Unreal Engine NeRF plug-ins \citep{lumaai_2023}.
Automated 3D environment synthesis holds great promise for film \textbf{virtual production} applications. For instance, combining environmental NeRF with light stages \citep{manzaneque2023revolutionizing} enables cost-effective scene shooting using {Volinga} suite \citep{volinga_2023}. Non-photorealistic stylized 3D assets and scenes are particularly beneficial for \textbf{animation production}, as demonstrated by the film {Elemental} \citep{hoffman2023creating, kanyuk2023singed}. 
Moreover, these techniques find possible applications in VR and video game development \citep{menapace2022playable}, enabling rapid stylization and editing of 3D scenes \citep{liu2023stylerf, fang2023text}.

\section{Open Challenges and Future Works}\label{sec:future}
From this survey, we identify under-explored problems and notable challenges in 3D neural stylization that are worth investigating in future work, which we discuss below. 

\paragraph{Generalization}
\noindent\textbf{Large-scale Scene Stylization.}
Most 3D neural stylization works focus on objects or object-centric scenes \citep{michel2022text2mesh, hertz2022spaghetti}, room-scale scenes \citep{pang2023locally, hollein2022stylemesh}, and outdoor inward-facing scenes \citep{huang2021learning, chiang2022stylizing}. Though \cite{kim2024fprf} extended novel view stylization to city scenes, it does not output stylized 3D assets. StyleCity (\citeyear{chen2024stylecity}) stylizes urban texture and sky but replies on heavy mesh representation.
3D assets and scenes can scale to as large as multi-room indoor scene \citep{straub2019replica, huang2022360roam}, architectural scenes \citep{martin2021nerf, wang2021ibrnet}, multi-block outdoor scenes \citep{tancik2022block, turki2023suds}, and even city-scale scenes \citep{xiangli2022bungeenerf, xu2023grid, li2023matrixcity}. These complex scenarios with intricate semantics are challenging for semantic alignment and computation efficiency.

%4D/Dynamic/Deformable/Time-varying Scene
\noindent \textbf{4D Scene Stylization.}
Limited literature exists on 4D scene stylization, with only a few notable works such as {SNeRF} (\citeyear{nguyen2022snerf}) and S-DyRF (\citeyear{li2024sdyrf}) presenting to stylize dynamic portraits and small scenes. Stylizing time-varying scenes with dynamic geometry and appearance changes \citep{park2021nerfies,song2023nerfplayer,yang2023gs4d}, or integrating time-related special effects \citep{shih2013data,logacheva2020deeplandscape}, poses significant challenges in maintaining temporospatial consistency in this domain.

\noindent \textbf{Generalizable Text-guided Stylization.} Text-guided 3D scene stylization and editing are still in the early stages. Data-driven generalizable text-guided 3D scene stylization or editing \textit{without re-training} is seldom explored yet \citep{fang2023text}, which demands more attention. It is worth investigation and exploration since current advanced large-language models such as BLIP \citep{li2023blip}, GPT \citep{brown2020language} enable infinite image-text pair data generation for data-driven model training.

\paragraph{Controllability}
\noindent\textbf{3D Reference-guided Stylization.}
Various modalities have been explored to be style references, especially images and text prompts. 3D to 3D geometric and appearance style transfer with 3D shape or 3D scene guidance is still underexplored \citep{yin20213dstylenet}. While 3D features can provide 3D-aligned holistic references for stylization, 3D feature extraction suffers from limited 3D datasets in a few categories.
With the rapid development of 2D-to-3D lifting techniques, there is potential to leverage large-scale pre-trained 2D models, such as 2D diffusion models \citep{rombach2022high, zhang2023adding}, as priors for context-aware scene stylization with 3D references. In addition, we expect more and more 3D pre-trained feature extractors and generative models with data in the wild to boost the 3D context-aligned style transfer. 

\noindent\textbf{Multi-modal Controls.} Currently, most research works focus on sole-reference guidance for stylization, while multi-modal reference can provide high accuracy and controllability on precise manipulation and design \citep{pang2023locally, bao2023sine, zhuang2024tip}. Therefore, it is worthwhile to explore the possibilities of joint supervision incorporating visual \citep{simonyan2014very}, textual \citep{radford2021learning}, semantic \citep{caron2021emerging}, and geometric features for 3D stylization.

\paragraph{Efficiency}
\noindent\textbf{Real-time Arbitrary Style Transfer of 3D Scenes.}
Modern photo or video filters support real-time processing \citep{ruder2016artistic, jamrivska2019stylizing}. For instance, \citet{ioannou2023neural} proposed a simplified style transfer architecture embedded into Unity rendering pipeline, enabling real-time depth-aware 2D style transfer. 
However, real-time stylizing a 3D scene given arbitrary styles remains challenging for some 3D representations due to the slow optimization process \citep{michel2022text2mesh, richardson2023texture, cao2023texfusion, yang20233dstyle} and slow rendering in neural fields \citep{chiang2022stylizing, zhang2022arf}. Even though there are some novel view stylization works \citep{li2019learning, liu2023stylerf, chen2024pnesm} that achieve arbitrary style transfer for speedy novel view synthesis, they fail to obtain stylized 3D scenes instantly. It is worth exploring to improve stylization speed by leveraging feed-forward networks \citep{aurand2022efficient}, 3D generative models \citep{cao2020psnet}, and advanced 3D representations such as 3DGS \citep{ liu2023stylegaussian, zhang2024stylizedgs}.

\paragraph{3D Consistency}
\textbf{Comprehensive View Planning for Complex Scenes.} Existing works have primarily focused on planning training views for object or room scenes for 3D stylization \citep{michel2022text2mesh, hwang2023text2scene, richardson2023texture, chen2023text2tex}, overlooking the crucial aspects of semantic- and instance-level view planning. Hence, it is a compelling research opportunity to investigate effective strategies for planning views in scenes characterized by intricate semantics such as cityscapes and multi-room scenarios.

\noindent\textbf{3D-Holistic Style Feature of Scenes.} The majority of works reviewed are supervised by large-data 2D pixel-level features extracted from multi-views \citep{michel2022text2mesh, zhang2022arf, kim2019transport}, since large-data 3D pre-trained models are still rare and expensive. Even though some works try to lift 2D content features to 3D before 3D stylization \citep{huang2021learning, liu2023stylerf, huang2022stylizednerf}, they still use view-dependent style features for final 3D stylization supervision. It is also impractical to lift 2D features to 3D at every iteration.  Some works supervise stylization with a 3D-aware style feature by averaging features of several views for a small object \citep{michel2022text2mesh, ma2023x}, which is not implementable for more views with limited memory.  
Per-view or multi-view supervision may not be sufficient to represent the whole 3D scene style feature, and worse may dilute the current single-view style from other views with conflicting gradients \citep{gao2023textdeformer}. More research and investigation are needed for efficient 3D-aware and even 3D-holistic style features for 3D stylization.

\paragraph{Evaluation}
\noindent\textbf{Standardized Evaluation Across Modalities.} The current evaluation metrics do not always align with human preference. User study is still widely adopted but it also prevents a precise quantitative analysis of method performance. The heterogeneity of datasets of different modalities also imposes great challenges for a fair and comprehensive comparison of works on different modalities. We believe there should be some variations from the evaluation of our benchmark, while concerns should be similar to criteria in Sec. \ref{sec:eval_metrics}.

\section{Conclusion}\label{sec:conclusion}
The report has explored the advancements in neural stylization techniques for diverse 3D data, including mesh, volume, neural fields, point cloud, and implicit shapes. Through this comprehensive survey of 3D neural stylization techniques and corresponding applications, we highlighted the importance of neural stylization in accelerating the creative process, enabling fine-grained control over stylization, and enhancing artistic expression in various domains such as movie making, virtual production, and video game development. Furthermore, we have introduced a taxonomy for neural stylization, providing a framework for categorizing new works in the neural stylization field. Our analysis and discussion of advanced techniques underscored the ongoing research efforts aimed at addressing limitations and pushing the boundaries of neural stylization in the 3D digital domain. In addition, we proposed a benchmark of 3D neural stylization, with which we aim to offer reference and inspiration for future 3D stylization works. Finally, we introduced practical applications and discussed open challenges and future works of 3D neural stylization.

\subsection*{Availability of data and materials} 
The experimental data for the benchmark of 3D stylization are available within the paper (Table~\ref{tab:datasets}). The evaluation codes are available via the open repository: \url{https://github.com/chenyingshu/advances_3d_neural_stylization}.

\bibliography{bibliography}% common bib file

\end{document}

%% file: fig_structure.tex
\begin{figure*}[!t]
\centering
\includegraphics[width=0.99\textwidth]{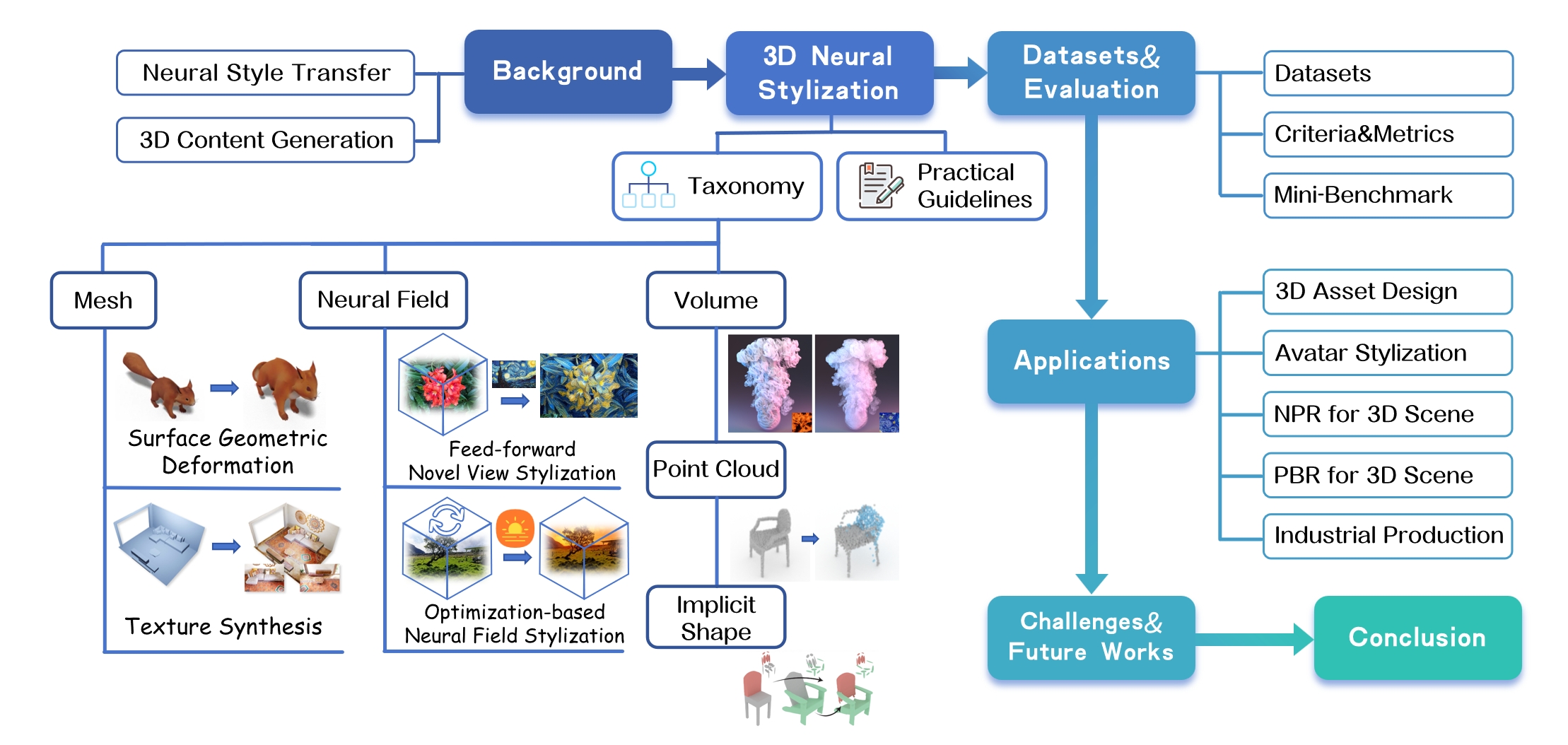}
\put(-433, 192){\footnotesize{\S\ref{sec:2d-image-guide}}}
\put(-433, 173){\footnotesize{\S\ref{sec:3dgen}}}
\put(-291, 160){\footnotesize{\S\ref{sec:3d_style_tax}}}
\put(-402, 128){\footnotesize{\S\ref{sec:3d_style_mesh}}}
\put(-313, 128){\footnotesize{\S\ref{sec:3d_style_neural_field}}}
\put(-223, 128){\footnotesize{\S\ref{sec:3d_style_volume}}}
\put(-223, 81){\footnotesize{\S\ref{sec:3d_style_point}}}
\put(-223, 38){\footnotesize{\S\ref{sec:3d_style_implicit_shape}}}
\put(-185, 145){\footnotesize{\S\ref{sec:practical}}}
\put(-135, 174){\footnotesize{\S\ref{sec:eva}}}
\put(-135, 92){\footnotesize{\S\ref{sec:app}}}
\put(-139, 14){\footnotesize{\S\ref{sec:future}}}

    \vspace{-5pt}
\caption{Structure of our survey. }
\label{fig_structure}
    \vspace{-10pt}
\end{figure*}